\documentclass[acmtog]{acmart}
\acmSubmissionID{180}  
\setcitestyle{square}
\pdfoutput=1

\acmJournal{TOG}
\acmYear{2018}\acmVolume{37}\acmNumber{6}\acmArticle{209}\acmMonth{11} \acmDOI{10.1145/3272127.3275027}

\setcopyright{none}

\usepackage[ruled]{algorithm2e} 
\usepackage{color}
\usepackage{steinmetz}
\usepackage{graphicx}
\usepackage{makecell}
\usepackage{amssymb}
\usepackage{amsmath}
\usepackage{mathtools}
\usepackage{multirow}
\usepackage{wrapfig}
\usepackage{algorithmic}
\usepackage{newlfont} 
\usepackage[english]{babel} 
\usepackage{enumitem}
\usepackage{microtype}
\usepackage{floatflt}
\usepackage{booktabs} 
\usepackage{subcaption}
\usepackage{epsfig}
\usepackage{xcolor}
\usepackage{bbm}
\usepackage{hyperref}

\newcommand{\bc}{\mathbf{c}}
\newcommand{\bdd}{\mathbf{d}}
\newcommand{\be}{\mathbf{e}}
\newcommand{\bff}{\mathbf{f}}

\newcommand{\bh}{\mathbf{h}}

\newcommand{\bs}{\mathbf{s}}
\newcommand{\bt}{\mathbf{t}}
\newcommand{\bu}{\mathbf{u}}

\newcommand{\bx}{\mathbf{x}}
\newcommand{\by}{\mathbf{y}}
\newcommand{\bz}{\mathbf{z}}

\newcommand{\bI}{\mathbf{I}}

\newcommand{\bR}{\mathbf{R}}

\newcommand{\bW}{\mathbf{W}}

\newcommand{\mS}{\mathcal{S}}

\newcommand{\mT}{\mathcal{T}}
\newcommand{\mM}{\mathcal{M}}
\newcommand{\mN}{\mathcal{N}}

\def\rev#1{{#1}}

\makeatletter
\def\@copyrightspace{\relax}
\makeatother

\definecolor{mygreen}{rgb}{0.3,0.9,0.4}
\newcommand{\eric}[1]{\textcolor{mygreen}{\bf [Eric: #1]}}
\newcommand{\vangelis}[1]{\textcolor{orange}{\bf [Vangelis: #1]}}
\newcommand{\hao}[1]{\textcolor{red}{\bf [Hao: #1]}}
\newcommand{\haibin}[1]{\textcolor{blue}{\bf [Haibin: #1]}}

\begin{document}

\title{Deep Part Induction from Articulated Object Pairs}

\author{Li Yi}
\affiliation{
  \institution{Stanford University}
}

\author{Haibin Huang}
\affiliation{
  \institution{Megvii (Face++) Research}
}

\author{Difan Liu}
\affiliation{
  \institution{University of Massachusetts Amherst}
}

\author{Evangelos Kalogerakis}
\affiliation{
  \institution{University of Massachusetts Amherst}
}

\author{Hao Su}
\affiliation{
  \institution{University of California, San Diego}
}

\author{Leonidas Guibas}
\affiliation{
  \institution{Stanford University}
}

\renewcommand{\shortauthors}{Yi et. al.}

\begin{abstract}
\rev{
Object functionality is often expressed through part articulation -- as when the two rigid parts of a scissor pivot against each other to perform the cutting function. Such articulations are often similar across objects within the same functional category.  In this paper we explore how the observation of different articulation states provides evidence for part structure and motion of 3D objects. Our method takes as input a pair of unsegmented   shapes representing two different articulation states of two functionally related objects, and induces their common parts along with their underlying rigid motion. This is a challenging setting, as we assume no prior shape structure, no prior shape category information, no consistent shape orientation, the articulation states may belong to objects of different geometry, plus 
we allow inputs to be noisy and partial scans, or point clouds lifted from RGB\ images. Our method learns a neural network architecture with three modules that respectively propose correspondences, estimate 3D deformation flows, and perform segmentation. To achieve optimal performance, our architecture alternates between correspondence, deformation flow, and segmentation prediction iteratively in an ICP-like fashion. Our results demonstrate that our method significantly outperforms state-of-the-art techniques in the task of discovering articulated parts of objects. In addition, our part induction is object-class agnostic and successfully generalizes to new and unseen objects.}
\end{abstract}

%
%


%
%


\begin{teaserfigure}
    \centering
    \includegraphics[width=\linewidth]{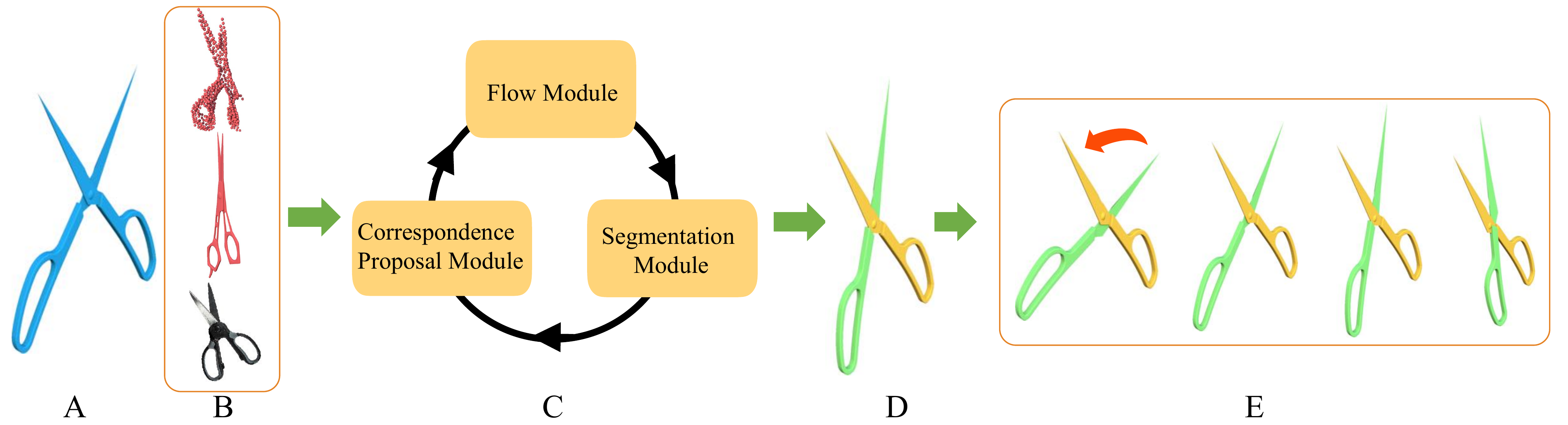}
    \caption{We present a neural network for discovering articulated parts of objects. Given a point set representing a 3D CAD model (A) and other functionally similar objects in the form of a scan, 3D mesh or RGB image (B), our network (C) can co-segment the input objects into their articulated parts and parse their underlying motion (D). The learned articulation can further be used for automatically animating the input CAD model (E).
    }
    \label{fig:teaser}
\end{teaserfigure}

\maketitle

\section{Introduction}

\rev{Our everyday living environments are largely populated with dynamic and articulated objects, which we can interact with through their moving parts e.g., swivel chairs, laptops, bikes, tools,  to name a few. In order for autonomous agents to correctly interact with such objects, the agents need to be equipped with algorithms that are able to parse these objects into their functional parts and motion. Decomposing 3D shape representations into their moving parts is also important for several graphics, vision, and robotic applications, such as predicting object functionality, human-object interactions, guiding shape edits, animation, and reconstruction.} 

Recently, with the availability of large 3D datasets and the use of deep learning techniques, significant progress has been made in the task of supervised part segmentation. Given a large volume of 3D shapes with part segmentation annotations, we can train deep neural networks to reliably segment new shapes from the same object category.
Although progress in supervised part segmentation is impressive, contemporary algorithms are still far inferior to humans when it comes to parsing 3D shapes from \emph{novel} object categories, \rev{and also to discovering new functional parts  whose types are not covered in the training sets. Being able to parse 3D objects into functional parts not seen before is fundamentally crucial towards building intelligent agents that must understand the object functionality, so as to have physical interactions with them, simulate such interactions, assist humans in augmented reality scenarios or autonomous robotics settings. Ideally, as new objects continuously emerge, an agent should possess the ability to induce their structures from the few observations and limited interaction experience.}

In this work, we are interested in discovering 3D object structure according to the \rev{mobility} of their underlying parts. Articulation can be a crucial clue in part structure determination, as it invariably involves motion of one part against others. We also aim to induce part structure from observations of different articulation states of an object in noisy settings e.g., 3D partial scans,  or even when the articulation state observations originate from geometrically different, yet functionally related, objects. Take Figure~\ref{fig:teaser} as an example: by observing different scissors under various opening angles, we expect to induce that a scissor is made of two thin blades that can rotate around a pivot. 

Mobility-based shape parsing brings a novel angle to the part determination problem. Among different principles for finding parts of objects (e.g., Gestalt theory~\cite{Palmer:1977}), the mobility-based principle provides an unambiguous decomposition.
Besides, mobility based parsing of an object can facilitate its functional understanding -- man-made objects are designed to function or interact with other objects (including humans) in ways realized by particular moving structures. As in the previous example, a scissor is designed with two pivoting blades to enable cutting. Recently, there have been increasing efforts devoted to obtaining a functional understanding of 3D objects from a motion perspective ~\cite{Pirk:2017, hu2017learning}. The recent work of ~\cite{hu2017learning} to infer part mobility types such as rotation and translation is particularly relevant to our effort, with the difference that they use a pre-segmentation of the object -- something we do not assume. 



\rev{Though seemingly trivial for humans, automatic part induction from observations of articulation states of objects is challenging for a number of reasons. First, the input objects can differ considerably in both geometry and pose. Second, the articulation differences must be  aggregated into coherently moving parts with clean boundaries. Third, in the case of  scans, part induction must be robust to both noise and missing data. }
   
We designed a deep neural network-based system to address the problem, encouraged by its robustness to data variation in various 2D and 3D data understanding tasks. However, unlike previous supervised methods, we do not assume any prior knowledge of the input shape class or object structure (e.g., part labels, or pre-defined components). \rev{The design of our neural network is motivated by the following  observations: (a) Establishing local correspondences between the input shapes requires much less global and high-level information compared with semantic understanding tasks such as classification and segmentation, thus a learned correspondence module is potentially much more transferrable to novel categories. Based on the predicted correspondences, strong cues can be obtained to infer a deformation flow field, capturing the differences in the articulation states of the two input shapes, (b) However, local correspondences are often ambiguous and fuzzy due to shape symmetries,  noise in scans, and geometric differences between the input shapes, thus one needs to incorporate global shape information to robustly translate correspondence cues into deformation flow. (c) Articulated parts can then be discovered by aggregating the deformation flow into rigid part motions. Following these observations, we designed a neural network that operates in three stages: (a) It learns to extract discriminative local features and propose possible  correspondences between the input shape pair, (b) learns to disambiguate correspondence confusion caused by symmetries, noise, and geometric differences by integrating global shape information to predict deformation flow, and finally, (c) learns to group points into parts from the predicted flow, leveraging a part rigidity assumption. These three stages are executed by neural network modules trained from a massive synthetic training dataset of articulated shapes within a multi-task learning framework. }

\rev{The three modules can also be executed iteratively to reinforce each other. This iterative procedure} is akin to ICP approaches or RANSAC-based algorithms ~\cite{Fischler:1981:RSC} that iterate between model fitting and geometric verification to discover primitives out of clean and same instance pairs. However, our learning based algorithm enjoys much stronger robustness to input data variation and corruption, which allows us to do induction from different instances that may have limited overlap, and avoids the tedious and error-prone tuning of sensitive hyper-parameters. 

\rev{We performed extensive qualitative and quantitative evaluation on both synthetic and real datasets. We also conducted ablation studies to confirm the utility of all the above mentioned network stages.  As we demonstrate in the results, previous methods largely fail to obtain satisfactory results even for objects with a single Degree of Freedom (DoF) in their joints, while our network successfully parses objects with either one or several DoFs. Overall, results demonstrate that our network dramatically outperforms existing state-of-the-art methods.}

\rev{In summary, this paper introduces a new deep learning method to parse 3D objects into moving parts based only on input static shape snapshots without any prior knowledge of the input object class or structure (part labels, or pre-existing components). Specifically, our method makes the following  key contributions}:

\begin{enumerate}
    \item Introduces a learning framework for mobility-based part segmentation from articulated object pairs that generalizes to novel object categories.
    \item \rev{Provides a new neural network modules for robust dense correspondence estimation between shapes with large geometric and articulation differences. The module is also capable of partial shape matching by inferring a correspondence mask that handles structural shape differences or missing data.}

    \item \rev{Provides a new neural network architecture, called PairNet, capable of inferring pairwise relationships between two input shapes. In our case, a PairNet infers deformation flow between the input shapes.}

    \item \rev{Provides a new neural network architecture for shape segmentation by generating hypotheses of rigid motions from deformation flow and sequentially extracting parts whose  motion is consistent with these hypotheses. The module implements a differential, neural-based  RANSAC procedure, which can be useful in other applications  requiring structure discovery from noisy observed data.}

    \item \rev{Demonstrates a neural net-based mutual reinforcement procedure iterating between correspondence, flow, and part estimation.}
\end{enumerate}

\section{Prior Work}
Our work is primarily related to  3D shape segmentation approaches that aim to extract rigidly moving parts from 3D meshes, point clouds or RGBD sequences. Our architecture extracts matching probabilities between points on 3D\ shapes at an intermediate stage, thus it is also related to learning-based 3D shape correspondence approaches.
We briefly overview these approaches
here.

\vspace{-0.15cm}
\paragraph{Rigid part extraction.} Given an input sequence of meshes, point clouds, or RGBD\  data representing an underlying articulated 3D object under continuously changing poses, various approaches have been proposed to detect and extract its rigidly moving parts.
 In contrast to all these approaches, we do not assume that we are given a continuous, ordered sequence of 3D\ object poses. In addition, our method can infer the rigidly moving parts of the input 3D\ object by matching it to other geometrically different shapes available in online repositories, or partial scans.
Thus, our setting is more general compared to previous rigid part extraction methods, yet we briefly overview them here for completeness.

In the case of RGBD sequences, early works attempt to estimate the 3D motion field (scene flow)
between consecutive frames
\cite{quiroga2014,Hornacek2014,vogel2015sfe,vogel2014view}. 
To recover parts, super segments can be extracted and grouped according to their estimated rigid transformations from the\ motion field \cite{golyanik2017multiframe}.
Alternatively, patches or points lifted from the RGBD frames can be clustered into segments based on their overall flow similarity across frames using Expectation-Maximization or coordinate descent formulations \cite{Stuckler:2015:EDR,jaimez15_mocoop}.
  Parts  can also be extracted from 3D\ point  flows
through direct clustering  on point trajectories \cite{Pillai2014articulatedmotions,Tzionas:ECCVw:2016}.
More similarly to our approach, the concurrent learning method by Shao et al. \shortcite{shao2018motion} trains a joint flow estimation and segmentation network for motion-based object detection in a scene. However, their approach mainly relies on RGB color to compute flow, and cannot handle complex structures or large articulation differences, as discussed in our results section.

In a similar spirit, in the case of raw 3D\ point cloud sequences, given established point-wise correspondences between consecutive point sets aligned through ICP, the point trajectories can be grouped through clustering and graph cut techniques \cite{Qing2016,Kim2016SimultaneousSE}. Alternatively, joints with their associated transformations can be fitted according to these trajectories based on consensus voting techniques, such as RANSAC \cite{Li:2016:MFU}. Another set of methods attempts to fit pre-defined skeletons or templates to the input sequences using non-rigid registration techniques (see \cite{dgp:2012:star} for a survey), random forest regressors and classifiers \cite{Shotton:2013:EHP}, or more recently through neural networks \cite{Toshev:2014:DHP,NewellYD16,Bogo:ECCV:2016,VNect2017,TomeRA17}. However, these methods are specific to particular classes of objects, predominantly human bodies. Finally, in the case of deforming meshes with explicit vertices and triangle correspondences, mean shift clustering on rotation representations can be used to recover the rigid parts of the deforming shape \cite{James:2005:SMA}. 

The above approaches decouple shape correspondences and part extraction in separate steps, which often contain several hand-tuned parameters. Our method instead computes point correspondences and part segmentations in a single deep-learned architecture trained from a massive dataset of shapes.

\vspace{-0.15cm}
\paragraph{Shape segmentation.} A widely adopted approach in 3D shape segmentation is to train a classifier that labels points, faces, or patches based on an input training dataset of shapes with annotated parts (see \cite{Xu:2016:DSA} for a recent survey). More recent supervised learning approaches employ deep neural net architectures operating on multiple views \cite{Kalogerakis:2017:ShapePFCN}, volumetric grids  \cite{maturana2015voxnets}, spatial data structures (kd-trees, octrees) \cite{wang2017ocnn,riegler2017octnet,klokov2017escape},
point sets \cite{qi2017pointnet,qi2017pointnetpp,su18splatnet}, surface embeddings \cite{Maron2017CNN}, or graph-based representations of shapes 
\cite{yi2017syncspeccnn}. These methods can only
extract parts whose labels have been observed in the training set, and cannot discover new parts. 

Our work is more related to co-segmentation and joint segmentation approaches that aim to discover common parts in an input set of 3D shapes without any explicit tags. To discover common parts, geometric descriptors can be extracted per point or patch on the input 3D shapes, or geometric distances can be computed between candidate shape segmentation, then clustering can reveal common parts of shapes \cite{sidi11cosegmentation,Hu:2012:CSS,vanKaick:2013:CAS}.
However, the resulting clusters cannot be guaranteed to correspond to common functional parts.
Our architecture is instead optimized to segment shapes under the assumption that functional parts in articulated objects predominantly undergo rigid motions, which is often the case for several man-made objects.

Alternatively, a family of approaches builds point-wise correspondences or functional maps between shapes
\cite{Golovinskiy:2009:CSM, huang2008non,Kim:2013:LPT,Huang:2014:FMN}, then employ an optimization approach that attempts to find parts that maximize geometric part similarity, or additionally satisfy cyclic consistency constraints. These methods largely depend on the quality of the initial correspondences and maps, while part similarity often relies on hand-engineered geometric descriptors. When input shape parts undergo large rigid transformations, their optimization approach can easily get stuck in unsatisfactory minima. Our approach instead learns to jointly extract shape correspondence and parts even in cases of large motions.
Our experiments demonstrate significantly better results than co-segmentation approaches.

\vspace{-0.15cm}
\paragraph{Shape correspondences.}
Since our architecture extracts point-wise correspondence probabilities as an intermediate stage, our work is also related to learning-based methods for 3D shape correspondences. \rev{In the context of deformable human bodies,
deep learning architectures operating on intrinsic representations \cite{masci2015geodesic,Boscaini2015spectral,Monti2017}
have demonstrated excellent results}. However, these methods cannot handle 3D shapes with largely different topology or structure, due to the instability of their spectral domain.
Volumetric, view-based and point-based neural networks have been proposed to learn point-based descriptors for correspondences between structurally and geometrically different shapes \cite{zeng20163dmatch,Huang:2017:LMVCNN,qi2017pointnetpp}. Given  shape correspondences extracted by these methods, one could attempt to recover rigid parts using RANSAC or Hough voting techniques \cite{Li:2016:MFU,Mitra:2006:PAS} in a separate step. However, as we demonstrate in our results section, decoupling correspondences and part extraction yields significantly worse results compared to our architecture. 


\section{Overview}
\label{sec:overview}
\begin{figure*}
    \centering
    \includegraphics[width=\linewidth]{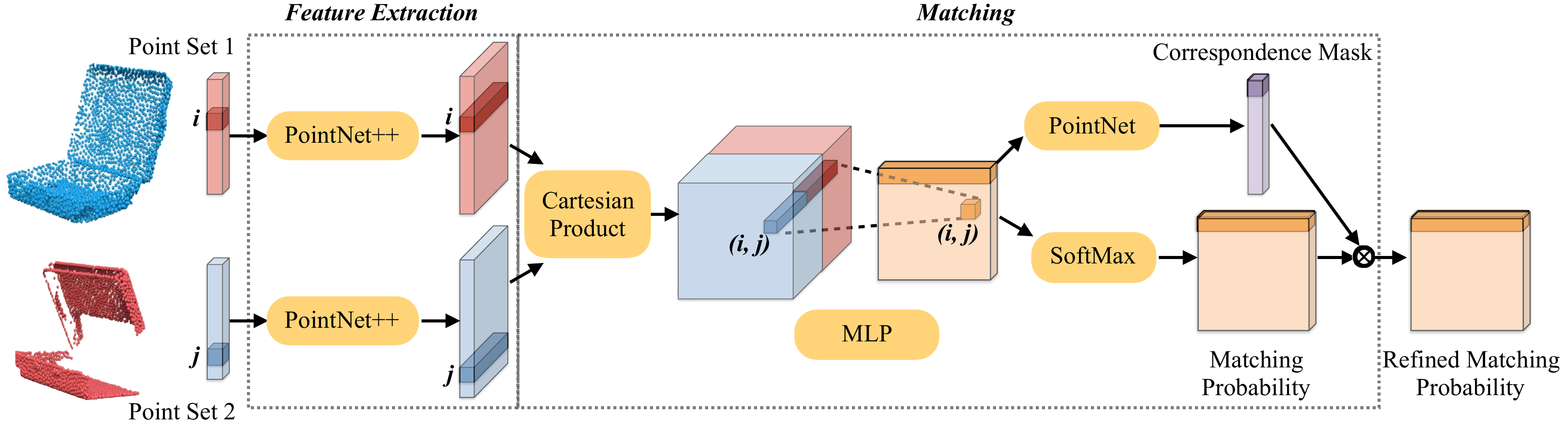}
    \vspace{-0.65cm}
    \caption{Correspondence proposal module. We use a PointNet++ based sub-module to extract point-wise features for the input  point clouds. The learned features are further fed into a matching sub-module for correspondence proposal. The sub-module also predicts a correspondence mask that determines which points should be matched or not.  }
    \label{fig:corrs_module}
    \vspace{-0.45cm}
\end{figure*}

\rev{Our method co-segments input 3D shapes into rigidly moving parts through a deep architecture shown in  Figure \ref{fig:teaser}(C).  Its modular design is motivated by the observation that estimating correspondences and deformation flows between shapes can provide cues for extracting rigidly moving parts, and in turn, the extracted piece-wise rigid motions can further improve the shape correspondences and deformation flows.}

\begin{figure*}
    \centering
    \includegraphics[width=\linewidth]{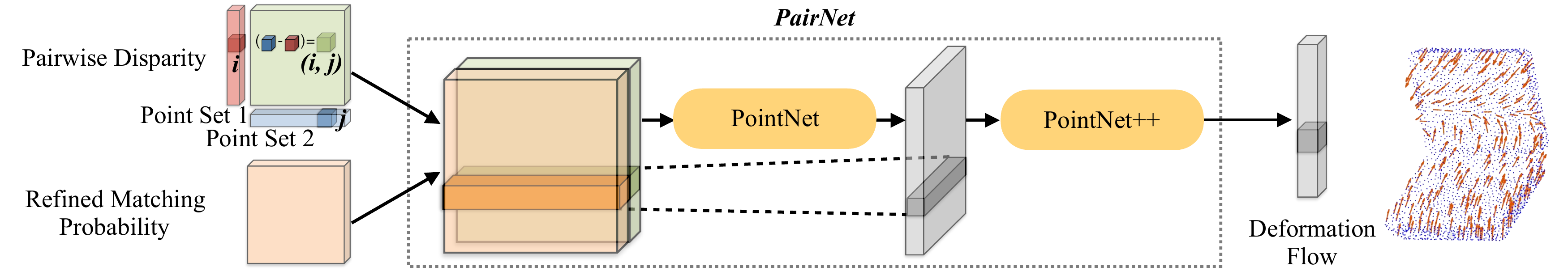}
    \vspace{-0.65cm}
    \caption{Flow Module. The refined matching probabilities are concatenated with the pairwise disparity and fed into the flow module. The flow module learns a point-wise deformation flow from one point set to the other.}
    \label{fig:flow_module}
    \vspace{-0.45cm}
\end{figure*}

\vspace{-0.15cm}
\paragraph{Neural network design.} \rev{In contrast to prior rigid part extraction and traditional ICP approaches, our shape correspondences are not based on closest points and hand-engineered geometric descriptors but instead are extracted through a learned neural network module}. This module, which we refer to as \emph{correspondence proposal module}, is trained to map the input shape pair geometry into probabilistic point-wise correspondences (Figure \ref{fig:corrs_module}). The module can handle large differences in both geometry and articulations in the input shapes as well as missing data and noise in the case of input 3D scans. Shape correspondences can provide strong cues for deformation flows and rigidly moving parts, but in general are not enough alone to reliably extract those. The reason is that correspondences are often ambiguous and fuzzy due to shape symmetries,  missing parts and noise in scans,  geometric and structural differences between the input shapes. Thus, our network incorporates another learned module, which we refer to as \emph{flow module}, that learns to robustly translate the extracted fuzzy correspondences into a  deformation flow field  (Figure \ref{fig:flow_module}). \rev{The module is based on a new type of network, called PairNet, designed to extract pairwise relationships between point sets.}
To discover the underlying shape structure, the deformation flows are aggregated into piecewise rigid motions that reveal the underlying shape parts. Instead of using hand-engineered voting or clustering strategies, the deformation flows are aggregated through a third, learned neural network module, called the \emph{segmentation module} (Figure \ref{fig:seg_module}). Since the number and  motion of parts are not known {\em a priori}, the  module first extracts rigid motion hypotheses from the deformation flows, discovers their support over the shape (i.e,  groups points that tend to follow the same underlying  motion), then sequentially extracts rigidly moving parts based on their support until no other parts can be discovered. \rev{The module is based on a new recurrent net-based architecture, called Recurrent Part Extraction Network, designed to handle sequential part discovery.}

\vspace{-0.15cm}
\paragraph{Iterative execution.} \rev{Inspired by ICP approaches which alternate between estimating shape correspondences and alignment, our architecture iteratively executes the correspondence, deformation flow, and segmentation modules in a closed loop}. Establishing correspondences provides cues for predicting deformation flow, the deformation flow helps extracting rigid parts, and in turn rigid parts helps improving shape correspondences and deformation flow. The loop is executed until the best possible alignment is achieved i.e., the total magnitude of the deformation flow field is minimized.  In practice, we observed that this strategy converges and yields significantly better segmentations compared to executing the network pipeline only once.

\section{Network architecture}
\label{sec:network}

\rev{Our network takes as input a pair of shapes $\{P,Q\}$ in the form of 3D point sets. If either  shape is in the form of a 3D mesh, we uniformly sample its surface using $N$ points ($N=512$ in our implementation). The only requirement for the input shape pair is that they should represent functionally related objects with rigidly moving parts in different articulation state. We do not make any assumptions on the shape orientation, order of points in the  point set representations, number of underlying parts and DoFs. Next, we discuss the modules of our architecture in detail.}

\vspace{-0.15cm}
\subsection{Correspondence Proposal Module}

The processing of the input shape pair $\{P,Q\}$ starts with the correspondence proposal module visualized in Figure \ref{fig:corrs_module}. Each shape in the pair is processed through a PointNet++ branch that outputs a $64-$dimensional feature representation for each point on the shape (we refer to the supplementary material for details regarding the PointNet++  structure). The two PointNet++ branches share their parameters i.e., have identical MLP layers such that the input geometry is processed in a consistent manner independently of the shape order in the pair.  Then, for each pair of points across the two shapes,  our architecture concatenates their extracted feature representation i.e., given the representation $\bu^{(p)}_i$ for a point $i$ on the shape $P$, and the representation $\bu^{(q)}_j$ for a point $j$ on the other shape $Q$, the resulting pair representation is $\bu_{ij} = \{\bu^{(p)}_i, \bu^{(q)}_j\}$. Performing this concatenation for all pairs of points (or in other words, using the cartesian product of the point set representations of the two shapes)
yields a tensor of size $N$x$N$x$2D$ ($N=512$ is the number of input points per shape,  $D=64$ in our implementation). Each pair representation $\bu_{ij}$ is transformed through a Multi-Layer Perceptron (MLP) (containing $128$ nodes in each of its $3$ hidden layers) into a confidence value $c_{i,j}$ that expresses how  likely the point $p_i$ matches, or corresponds to point  $q_j$. The confidence is further transformed into a probability using the softmax function. Executing the same MLP for all pairs of points and passing the resulting confidences through softmax yields a pairwise matching probability matrix $\mM$ with size $N$x$N$ that describes the probability of matching any pair of points on the two shapes. The network is then trained to output high probabilities for corresponding points in the training data.

Since there might be points on the first shape $P$ that have no correspondences to any other point on the second shape $Q$ due to missing data or structural differences, the correspondence proposal module also outputs the probability for each point on the first shape being matched  or not. \rev{We refer to this output as correspondence mask}. Specifically, the confidences of each point on the shape $P$, stored in the vector $\bc_i = c_{i,*}$ (where $*$ means all other points on the shape $Q$) are processed through a PointNet, that aggregates confidences throughout the whole vector, to determine the probability $c_i$ for matching the point $i$ of shape $P$ with any other point on the shape $Q$. During training, the network also receives supervisory signal for this output i.e., whether each training point possesses a correspondence or not. We multiply each row of the above pairwise matching probability matrix $\mM$ with the estimated probability of the corresponding point being matched, resulting in the refined pairwise matching probability matrix $\hat{\mM}$ as the output of our correspondence proposal module.

\vspace{-0.15cm}
\subsection{Flow Module}

The flow module, visualized in Figure~\ref{fig:flow_module}, aims to produce a 3D deformation flow field $\bff$ from the shape $P$ to shape $Q$, which  provides cues for determining common shape parts along with their rigid motions. One possibility would be to use the pairwise matching probability matrix $\hat{\mM}$ alone to infer this field. However, we found that this is not sufficient, which is not surprising since the flow should also depend on point positions as well i.e., if we rotate one shape, the pairwise correspondence probabilities should remain the same, yet the flow would change. Thus, we pass both point position and correspondence information as input to the flow module. 

Specifically,  for each pair of points across the two shapes, we compute their relative displacement, or disparity i.e., $\bdd_{ij} = \bx^{(q)}_j - \bx^{(p)}_i$, where $\bx^{(p)}_i$ is the position of point $i$ from shape $P$, and $\bx^{(q)}_j$ is the position of point $j$ from shape $Q$. Computing the all-pairs displacement matrix yields a $N$x$N$x$3$ pairwise displacement matrix, where $3$ corresponds to the $xyz$ channels.
The pairwise displacement matrix is concatenated together with the refined matching probability matrix $\hat{\mM}$ along their $3$rd dimension, forming a $N$x$N$x$4$ matrix passed as input to the flow module.

The flow module first processes each row of the stacked matrix through a PointNet (see supplementary material for architecture details). The PointNet aggregates information from all possible displacements and correspondences for each point on shape $P$. Note here that the set, which PointNet aggregates on, is the set of points from $Q$. The output of the PointNet is a representation of dimensionality $256$ that  encodes this aggregated information per point. Processing all points of shape $P$ through the same PointNet, yields a matrix of size $N$x$256$ that stores all point representations of shape $P$. This point set representation is then processed through a PointNet++ (see  supplementary material for architecture details). The PointNet++ hierarchically captures local dependencies in these point representations (e.g., neighboring points are expected to have similar flows), and outputs the predicted flow field $\bff$ ($N$x$3$) on shape $P$. As explained in the next section, the module is trained to extract flow using supervisory signal containing the ground-truth 3D flows for several possible shape orientations to ensure rotational invariance. 

We refer to the combination of the PointNet and PointNet++ as PairNet. PairNet takes a pairwise matrix between two sets as input. It first globally aggregates information along the second set through the PointNet to extract a per-point representation for each point in the first set. The point representations are then hierarchically aggregated into a higher-level representation through a PointNet++, to encode local dependencies in the first set.


Even facilitated by the power of neural networks, the deformation flow estimation is still far from perfect (Figure~\ref{fig:iter_vis}). It is inherently not well-defined due to geometric or structural differences across shape pairs. As described in the next paragraph, learning plays a key role to reliably extract parts from the estimated  flow.

\vspace{-0.15cm}
\subsection{Segmentation Module}
\label{sec:segmodule}

\begin{figure*}
    \centering
    \includegraphics[width=\linewidth]{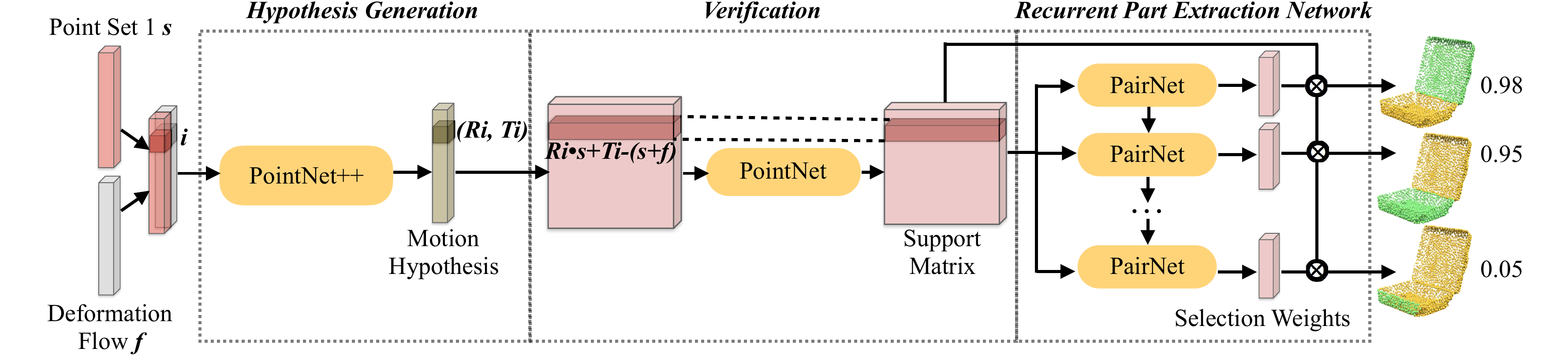}
    \vspace{-0.65cm}
    \caption{Segmentation module. The predicted deformation flow on the first point set together with its point positions are fed into this module, which acts as a neural net-based, differentiable sequential RANSAC . Similar to sequential RANSAC, the inputs are processed through three sub-modules including hypothesis generation, verification and \rev{recurrent part extraction network}, resulting in a set of soft segmentation indicator functions and part confidence scores.}
    \label{fig:seg_module}
    \vspace{-0.45cm}
\end{figure*}

Given the estimated deformation field $\bff$  from shape $P$ to $Q$, the segmentation module decodes it towards rigid motion modes as well as the corresponding part segments between the two shapes (Figure \ref{fig:seg_module}). The design of the module is inspired by  RANSAC\ approaches. First, the module generates hypotheses of rigid motions, then for each rigid motion it finds support regions on the shape (i.e., groups of shape points that follow the same rigid motion hypothesis), and finally extracts the rigid parts from the support regions one-by-one starting from the most dominant ones i.e., the ones with largest support. In contrast to traditional RANSAC\ approaches that employ hand-engineered techniques with user-defined thresholds in each of the stages (e.g., the target number of segments, inlier thresholds etc), our module implements them through learned neural network layers. In this sense, this module can be considered as a  neural network based, differentiable sequential RANSAC procedure. Different from \cite{brachmann2017dsac}, we learn to generate hypothesis efficiently to avoid expensive sampling, plus we have a sequential hypothesis selection step which allows decoding multiple modes from input samples. In the next paragraphs, we explain this module in detail.

\vspace{-0.15cm}
\paragraph{Hypothesis generation.} \rev{The first stage of our segmentation module is to generate hypotheses of  candidate part motions from the shape $P$ towards corresponding parts of the shape $Q$}. We here use a PointNet++ (see supplementary material for architecture details), to hierarchically aggregate the deformation flow $\bff$ along with the point positions of the shape $P$  to generate rigid motion hypotheses. The point positions are used as additional input to this stage along with the flow, since knowing the flow field alone without knowing the underlying geometry is not sufficient to determine a rigid motion. The output of the PointNet++ is a hypothesis for a rigid motion estimated per point $i$ of the shape $P$.
We experimented with various rigid motion output parameterizations, including predicting directly a $3$x$3$ rotation matrix and a $3$x$1$ translation vector, axis-angle and quaternion parameterizations of rotations, and affine matrice outputs followed by SVD to extract their rotational component. We found that the best performing rigid motion parameterization was through  a $3$x$3$ matrix $\hat{\bR}_i$ and a $3$x$1$ vector $\hat{\bt}_i$, from which the rotational component  is computed as $\bR_i = \hat{\bR}_i + \bI$ followed by an SVD to project the matrix to the nearest orthogonal matrix, while the translational component is computed after applying the inferred rotation: $\bt_i = -(\bR_i-I) \cdot \bx^{(p)}_i+ \bff_i + \hat{\bt}_i$, where $I$ is the identity transformation, $\bx^{(p)}_i$ is the $i^{th}$ point position and $\bff_i$ is the corresponding flow. We suspect that this parameterization resulted in better rigid motion and segmentation estimates
due to the fact that the rotational and translational components of a rigid motion are not independent of each other and also because the elements of $ \hat{\bR}_i$ and $ \hat{\bt}_i$  have more compatible scales. Thus, for computing rotations, we predict the intermediate matrix $\hat{\bR}_i$ which is equal to the zero matrix in case of the identity transformation (i.e., a ``residual'' rotation matrix), while the translation is predicted conditioned on the estimated rotation (again, a ``residual'' translation vector). We found that these residual representations are much easier to train and yield the best performance in terms of rigid motion and segmentation estimation. 

\vspace{-0.15cm}
\paragraph{Support prediction.} Following the rigid motion hypothesis generation stage, our segmentation module predicts a probability for each point on the shape $P$ to support, or in other words follow, each generated rigid motion hypothesis. To predict this probability, the segmentation module first examines how well each rigid motion hypothesis explains the predicted flow per point. This can be examined by applying the motion hypothesis to each point, computing the resulting displacement, and then comparing it with the predicted flow from our previous module. The displacement of a point $i'$ after applying the hypothesis $i$ is computed as $\bdd_{i,i'} = \bR_{i} \bx^{(p)}_{i'} +\bt_{i}-\bx^{(p)}_{i'}$, and the difference between this displacement and predicted flow is simply calculated as $\bdd_{i,i'} - \bff_{i'}$. Computing the flow difference for each point $i'$ and each motion hypothesis $i$ yields a $N$x$N$ pairwise flow difference matrix, where the rows correspond to rigid motion hypotheses (same number as the number of points of $P$) and columns corresponds to points. The module then aggregates information from all flow differences per rigid motion hypothesis to calculate its support. This is done through a PointNet operating on each row of the flow difference matrix. The PointNet is trained to output the probability of each point supporting a rigid motion hypothesis based on the estimated flow difference. Computing these probabilities for all available hypotheses yields a $N$x$N$ matrix, which we refer to as support matrix $\mS$. The rows of the matrix correspond to candidate rigid motion hypotheses and columns correspond to the per-point support probabilities.

\vspace{-0.15cm}
\paragraph{Rigid part extraction.}
The last stage of the segmentation module is to decode the support matrix into a set of rigid segments in a sequential manner.
\rev{Decoding is performed through a recurrent net-based architecture, that we refer to as Recurrent Part Extraction Network (RPEN). The RPEN outputs one segment at each step and also decides when to stop.
It maintains a hidden state $\bh_t=(\be_t, \bz_t, \bs_t)$, where $\be_t$ represents an internal memory encoding the already segmented regions so that subsequent steps decode the support matrix into different segments, $\bz_t$ is a  learned representation of the recurrent unit input designed to modulate the  support matrix such that already segmented regions are downplayed,
and $\bs_t$ denotes a learned weight representing the importance of each hypothesis for segment prediction.}

At each step $t$, the recurrent unit transforms its input support matrix $\mathcal{S}$, as well as the memory $\be_{t-1}$ from the previous time step, into a compact representation $\bz_t$ through a PairNet (same combination of PointNet and PointNet++ used in the flow module). Each row of $\mathcal{S}$ is concatenated with $\be_{t-1}$, forming a $N$x$N$x$2$ matrix, and is then fed to the PairNet to generate $\bz_t$ with dimensionality $N$x$32$. Then the recurrent net decodes the representation $\bz_t$ into the following two outputs through two different PointNets:
(a) a continuation score $r_t$ (a scalar value between $0$ or $1$) which indicates whether the network should continue predicting a new segment or stop, and (b) a $N$x$1$  vector $\bs_t$ representing how much weight the support region of each hypothesis should be given to determine the  output segment of shape $P$ at the step $t$.
The soft segmentation assignment variable $\by_t$ is generated by computing the weighted average of the corresponding hypothesis support probabilities, or mathematically $\by_t = \mS^T \cdot \bs_t $. At inference time, the per-point soft segment assignments are converted into hard assignments through graph cuts \cite{Boykov2001}.
\rev{The associated rigid motion is estimated by fitting a rigid transformation to the deformation flow $\bff$ on the segment. By applying the fitted rigid transformation on the points of the segmented part of shape $P$, we find the corresponding points on shape $Q$ using a nearest neighbor search, then execute the same graph cuts procedure to further refine the corresponding segmented part on $Q$.}

At each step, the Recurrent net also updates the hidden internal state $\bh_t$. The hidden state is updated in the following way: $\bz_t = f_{\text{PairNet}}(\mathcal{S}, \be_{t-1})$, $\bs_t = g(z_t)$, $\be_t = (1-\be_{t-1})\odot(\mS^T\bs_t)+\be_{t-1}$,
where $f_{\text{PairNet}}(\cdot)$ denotes the operation of PairNet, $g(\cdot)$ denotes a PointNet operation and $\odot$ denotes element-wise multiplication. Notice $\bs_t$ in the hidden state is directly outputted to determine the segment of shape $P$. The segmentation module stops producing segments when the continuation score falls below $0.5$ ($50\%$ probability).

\vspace{-0.15cm}
\subsection{Iterative Segmentation and Motion Estimation}

\begin{figure}
    \centering
    \includegraphics[width=\linewidth]{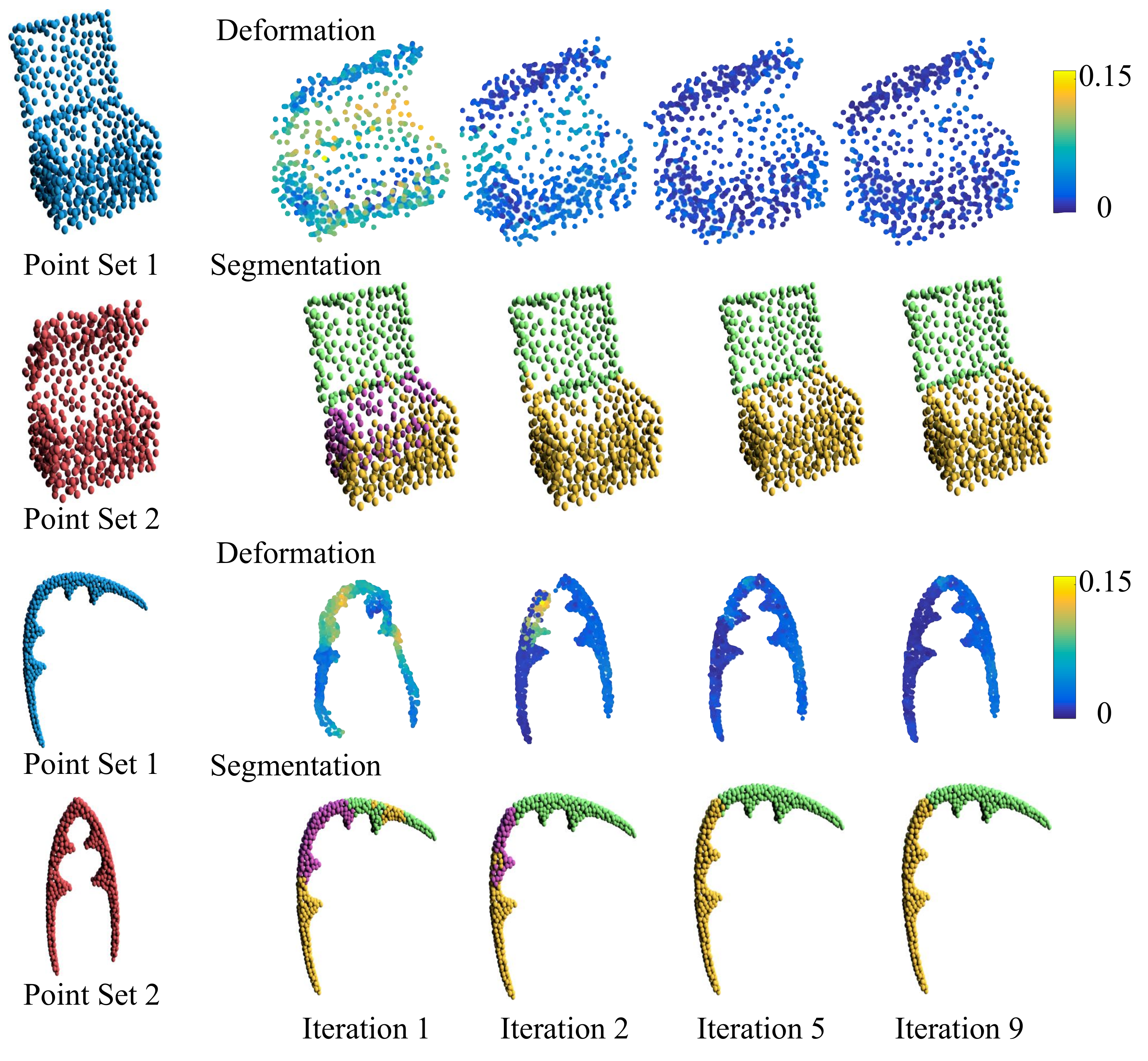}
    \vspace{-0.45cm}
    \caption{Iterative refinement of deformation flow and segmentation. The outputs usually converge after 5 iterations. }
    \label{fig:iter_vis}
    \vspace{-0.35cm}
\end{figure}

When there are large articulation differences between the two input shapes or in the presence of noise and outliers in input scans, the execution of a single forward pass through our architecture often results in a noisy segmentation (Figure \ref{fig:iter_vis}). In particular, an excessive number of small parts is often detected, which should  be grouped  instead into larger parts. We found that the main source of this problem is the estimation of the flow field $\bff$, which tends to be noisy in the above-mentioned conditions.

\rev{Inspired by ICP-like approaches, our method executes an iterative procedure to refine the prediction of the flow field}. Specifically, given an initial estimated flow field $\bff$, along with rigid motions $\{H_t\}$ and segmented parts $\{P_t\}$ of shape $P$ based on a first forward pass of the initial shape pair through our correspondence, flow, and segmentation modules, we produce a new deformed version $P'$ of the original shape $P$. \rev{The deformed shape is produced by applying the detected rigid motions on the associated parts of the original shape} i.e., $H_t \cdot P_t$. Then we compute a new ``residual'' flow field $\bff'$ by passing the pair $\{P',Q\}$ through the same correspondence and flow module of our network. The ``residual'' flow field is added to the piecewise rigid deformation field $P'-P$ to compute a new refined flow field i.e., $P'-P+\bff'$. This refined flow field is subsequently processed through our segmentation module to further update the rigid motions $\{H_t\}$ and segmented parts $\{P_t\}$ of the original shape $P$.
The procedure can repeat till it converges or reaches a maximum number of iterations ($10$ in our case).

\vspace{-0.15cm}
\paragraph{Initialization.} As in the case of ICP-like approaches, when the orientation of the two shapes differs significantly e.g., their upright or front-facing orientation is largely inconsistent, the algorithm might converge to a suboptimal configuration. In this case, at an initialization stage, we search over several different 3D global rotations for one of the two shapes (in our implementation, $48$ 3D rotations, uniformly sampled from SO(3)), and initialize our iterative procedure from the one that yields the smallest flow magnitude according to the flow module. The flow evaluation is fast: it takes about $10$ ms (measured on a TitanX GPU) to evaluate each candidate rotation, thus in the case of inconsistent shape orientation for the input pair, this initialization stage takes less than half of a second.

\vspace{-0.15cm}
\paragraph{Results.} The iterative version of our algorithm significantly improves both the deformation flow between $P$ and $Q$ as well as the segmentation of $P$, as demonstrated in the results section. We also refer to Figure \ref{fig:iter_vis} for  qualitative results.

\section{Training}
\label{sec:training}
Our network is trained on a large synthetic dataset consisting of pairs of shapes with ground-truth annotations of corresponding parts along with their rigid motions. \rev{In this section, we describe the synthetic training dataset, multi-task objective function, then we describe the training procedure.}

\vspace{-0.15cm}
\subsection{Training Dataset} Our training dataset is based on semantic part annotations \cite{yi2016scalable} of ShapeNetCore \cite{chang2015shapenet}, which contains $16,881$ segmented man-made shapes in 16 categories. \rev{For each shape $P$ in this dataset, we generate 2 deformed versions of it by applying  rotations on randomly picked segments} about random axes passing through 
the contact regions between adjacent parts, including random translations along axes that are perpendicular to  planar approximations of contact regions under the constraint that the resulting transformations keep the parts connected. We generate one pair of shapes for each $P$, resulting in a total number of $16,881$ pairs, with $2$ to $6$ moving parts per pair. We set a ratio of $1:3$ for the percentage of translations versus rotations in our deformation. For each pair of deformed shapes, we additionally generate 5 pairs of synthetic scans from random viewpoints and we conduct farthest point sampling on each synthetic scan, resulting $512$ points per scan, which are then normalized to have bounding box centered at $(0,0,0)$ and with a diagonal $1$. 

\rev{We refer the readers to the supplementary material for a list of the categories and example visualizations of the training set.} The resulting shape pairs contain (a) ground-truth part correspondences, (b) reference rigid motion per part, (c) ground-truth flow field from one shape to the other, (d) point-wise correspondences of corresponding parts since we know the underlying rigid motion that maps the points of one part onto the other, and finally (e) a binary flag for each point indicating whether it has correspondences with any other point on the other shape or not.

\vspace{-0.15cm}
\subsection{Multi-task objective function}
Given a training set $\mT$ of shape pairs, \rev{the network is trained according to a multi-task objective such that the  predictions of each module in our architecture agrees as much as possible with the ground-truth annotations}. Specifically, we minimize a loss $L$ that includes terms related to the correct prediction of point correspondences along with existence of those $(L_{corr})$,  flow field  $(L_{flow})$, rigid motions $(L_{motion})$, and part segmentations $(L_{seg})$:
\begin{eqnarray}
L = \sum\limits_{ \{P,Q\} \in \mT} & \Big( L_{corr}(P,Q) + L_{flow}(P,Q) + \\
& L_{motion}(P,Q) + L_{seg}(P,Q) \Big)
\end{eqnarray}
We discuss the above loss terms in the following paragraphs.

\vspace{-0.15cm}
\paragraph{Correspondence loss.} Given a set $\mM$ of ground-truth pairs of corresponding points across a training shape pair  $\{P,Q\}$, and a set $\mN$ of points on shape $P$ that do not match any point on shape $Q$, the correspondence loss penalizes outputs of the correspondence module that are incompatible with the above sets. Specifically, the correspondence loss is expressed as a weighted sum of two losses. \rev{The first loss $L_a$ penalizes low probabilities for matching point pairs deemed as corresponding in the set  $\mM$. The second loss $L_b$  penalizes low probabilities for  matching a point that does not belong to the set $\mN$ and similarly low probabilities for  not matching a point that belongs to the set $\mN$.} Since the matching of point pairs in our correspondence module is posed as a multi-class classification problem, $L_a$ is formulated as a softmax classification loss. Since the decision of whether a point on shape $P$ has a correspondence  is a binary classification problem, $L_b$ is expressed through binary cross-entropy. Specifically, the correspondence loss is set as $L_{corr} = \lambda_a L_a + \lambda_b L_b$, where:
\begin{align*}
& L_a = -\sum\limits_{ \{p,q\} \in \mM} \log( softmax(c_{p,q})) \\
& L_b = - \sum\limits_{p \notin \mN} \log( sigmoid(c_{p} ) ) - \sum\limits_{p \in \mN} \log( 1-sigmoid(c_{p})) \end{align*}
and $\lambda_a$, $\lambda_b$ are weights both set to $1$ through hold-out validation.

\vspace{-0.15cm}
\paragraph{Flow loss.} Given the ground truth flow field $\bff^{(gt)}(P,Q)$  and the predicted field  $\bff(P,Q)$ from our flow module for a training shape pair $\{P,Q\}$, the flow loss directly penalizes their difference using the $L^2$ norm: $L_{flow}(P,Q) = \lambda_{c}|| \bff^{(gt)} - \bff||^2$,
where  $\lambda_c$ is a weight for this loss term again set to $1$ through hold-out validation.

\vspace{-0.15cm}
\paragraph{Rigid motion loss.} The rigid motion loss penalizes discrepancies between the ground-truth rotations and translations assigned to points of  rigid parts of the training shapes and the hypothesized ones. We found that using a loss operating directly on the elements of the rotational and translational component of the rigid motion (e.g., Frobenius or $L^2$ norm difference) resulted in poor performance in terms of flow and segmentation prediction. The reason was that it was hard to balance the weights between the two components i.e., often either the rotational or translational component\ dominated at the expense of the other. 

We instead found that significantly better performance was achieved when using a loss that compared the positions of  points belonging to the same underlying rigid part of shape $P$, after applying the hypothesized rigid motion,  with the positions of corresponding points on shape $Q$. Specifically, for each pair of ground-truth corresponding points in the set $\mM$ for the training shape pair  $\{P,Q\}$, we find all other pairs of corresponding points  belonging to the same underlying rigid part, and measure the $L^2$ norm of the difference in the positions of points on shape $Q$ and position  of the corresponding points on shape $P$ after applying each hypothesized rigid motion:


\begin{align*}
L_{motion} = \lambda_d \sum\limits_{ \{p,q\} \in \mM \,\,}
\sum\limits_{ \substack{ \{p',q'\} \in \mM \\ part(p')=part(p) } }
\!\!\!\!\!\!\!\! || q' -( \bR_p p' + \bt_p) ||^2
\end{align*}
where  $\lambda_d$ is a weight for this loss term set to $1$ through
hold-out validation, and $part(p),part(p')$ return the part that the points $p,p'$ belong to respectively i.e., in the above summation, we consider pairs of $\{p,p'\}$ belonging to the same rigidly moving part.



\vspace{-0.15cm}
\paragraph{Segmentation loss.} We designed the segmentation loss such that supervisory signal is received on  (a) the support prediction stage of the segmentation module, which outputs the support matrix $\mS$ storing the probability of each point on a shape to follow the rigid motion hypotheses generated from each other point on the shape, (b) the recurrent net stage of the segmentation module, which outputs the soft segmentation assignment variables $\by_{t}$ and the continuation score $r_{t}$ at each step $t$. We found that incorporating supervision for both these stages of the segmentation module offered the best performance. 
Specifically, the segmentation loss is a weighted sum of three terms: $L_{seg} = \lambda_e L_e + \lambda_f L_f + \lambda_g L_g  $, where $L_e$ is a loss term that evaluates the predicted support matrix,  $L_f$ evaluates the assignments of the recurrent net segmentation  variables,  $L_g$ evaluates the recurrent net continuation scores, and $\lambda_e,\lambda_f,\lambda_g$ are loss term weights set to $0.5,1,1$ respectively through hold-out validation.
 
The prediction of the support matrix $\mS$ can be treated as a binary classification problem: a point $p$ on shape $P$ either follows or not  the rigid motion hypothesis generated from another point $p'$ on the same shape. Since we know whether $\{p',p\}$ fall onto the same rigid part or not based on the ground-truth annotations, we can evaluate the predicted support matrix through binary cross-entropy:
\begin{equation*}
L_s = - \sum\limits_{ \substack{p,p':\\part(p)=part(p')} } \log( S_{p,p'} ) - \sum\limits_{\substack{p,p':\\part(p) \ne part(p')} } \log( 1-S_{p,p'})
\end{equation*}
where $part(p), part(p')$ return the rigid part the points $p,p'$ belong to respectively.

Evaluating the output segmentation assignment variables of the recurrent net is more challenging because the order of the output segments does not necessarily match with the order of ground-truth segments specified for the training shapes. To handle the uncertainty in the order of the output segments, we use a loss function inspired by Romera and Torr \shortcite{romera2016recurrent}.
Assuming that the shape $P$ in an input training  pair has $K$ annotated rigid segments, the segments can be represented through binary indicator vectors $\{\hat{\by}_k\}_{k=1...K}$, where each vector $\hat{\by}_k$  stores a binary value per point indicating whether it belongs to the segment with index $k$ or not. The output of our recurrent net is  a soft indicator vector $\{\by_t\}_{t=1...T}$, which contains the probability of a point belonging to the output segment at step $t$ (where $T$ is the number of the executed RNN steps). During training, we set the maximum number of RNN execution steps as $T=K+2$ i.e., we predict two extra segments compared to ground-truth (we experimented with more steps, but did not have any noticeable effect in performance). We use the Hungarian algorithm \cite{kuhn1955hungarian} to find a bipartite matching between the ground-truth segment indicator vectors $\{\hat{\by}_k\}$ and predictions $\{\by_t\}$,  then employ a relaxed version of the Intersection over Union (IoU) score \cite{krahenbuhl2013parameter} to evaluate the matched pairs of segments. The relaxed IoU between a predicted segment output $\by_t$ and a matched ground-truth  segment $\hat{\by}_{k(t)}$ is defined as: $IoU(\by_t, \hat{\by}_{k(t)})=\frac{<\by_t, \hat{\by}_{k(t)}>}{||\by_t||_1+|| \hat{\by}_{k(t)}||_1-<\by_t,\hat{\by}_{k(t)}>}$. Then the loss term $L_f$ is expressed as the negative of a sum of IoUs over $K$ matched segment pairs: $L_f = -\sum\limits_{t=1...K} IoU(\by_t, \hat{\by}_{k(t)})$

Finally, the loss term $L_g$ evaluates the recurrent net continuation scores, penalizing low continuation probability for the first $K-1$ RNN\ execution steps, and high continuation probability after performing $K$ steps. The decision to continue producing segments can be considered as a binary classification problem, thus the loss term on continuation can be expressed through binary cross-entropy: $L_g = \sum\limits_{t < K} \log(r_t)+ \sum\limits_{t \ge K} \log(1-r_t)$
 
\vspace{-0.15cm}
\subsection{Training Procedure and Implementation Details} We minimize our loss function using the Adam variant of batch gradient descent with a learning rate of $0.0001$. To better balance variant losses, we adopt a stage-wise training strategy.
\rev{We first optimize the correspondence proposal and flow module by minimizing the sum of $L_{corrs}$ and $L_{flow}$ for $100$ epochs. Then we feed the ground truth flow to the segmentation module and optimize the hypothesis generation and verification submodule by minimizing $L_{motion}$ and $L_e$ for $100$ epochs. Finally we include all the loss terms in the optimization and further train the whole pipeline in an end-to-end manner for another $100$ epochs with a learning rate decay set to a factor of $0.001$.}

\vspace{-0.15cm}
\rev{\paragraph{Hyper-parameter selection} Our architecture makes extensive use of PointNet and PointNet++ networks. Their number and type of layers were selected using the default architecture blocks provided in \cite{qi2017pointnet,qi2017pointnetpp}. Regarding the layer hyper-parameters of our architecture (grouping radius in PointNet++,  dimensionality of intermediate feature representations, memory size in our RPEN), we performed a grid search over different values in a hold-out validation set with ground-truth shape segmentations, and selected the ones that offered the best performance in terms  of IoU.
}

\vspace{-0.15cm}
\rev{ \paragraph{Implementation} Our method is implemented in Tensorflow. Our source code, datasets and trained models are available in our project page:} \url{https://github.com/ericyi/articulated-part-induction}

\section{Experiments}
\label{sec:evl}
\rev{In this section, we evaluate the quality of our approach and compare it to state-of-the-art methods.} We conducted experiments on both synthetic and real datasets and demonstrate the performance of the whole framework as well as each module component.

\vspace{-0.15cm}
\subsection{Test Dataset}
\paragraph{Synthetic Dataset.}  We leveraged the annotated dataset in \cite{hu2017learning}, which contains articulated 3D CAD models with ground truth part segmentations and motion annotations, and generated three synthetic datasets: 1) Point cloud pairs originating from two different articulations of the same 3D CAD model (\textbf{SF2F});  2) Point cloud pairs consisting of one full shape and one partial scan from the same 3D model but with different articulations(\textbf{SF2P}); 3) RGBD pairs consisting of two partial views of the same object with different articulations(\textbf{SP2P}). For each CAD model, we randomly transformed its moving parts 10 times following the part segmentation and motion annotations in \cite{hu2017learning}. \rev{Then we randomly create 5 shape pairs by randomly selecting two  different articulations of the same model out of its $10$ generated configurations.  We also added random perturbations to the global poses of the shapes}. Then we uniformly sample points from the full shapes or conduct virtual scans from random viewpoints to generate partial point clouds using rendering tools developed from \cite{hassner2015effective}. We normalize the point sets so that their bounding boxes are centered at $(0,0,0)$ and have diagonals with length 1. We removed categories whose part motions cannot be distinguished from sampled point clouds. This could be due to either tiny parts (e.g., the button on a remote control, which might not be sampled in the point cloud, or sampled by less than 10 points), or due to rotational symmetries of a part (a rotating bottle cap cannot be distinguished in the sampled point clouds, because its rotations yield almost the same points). In total, for each synthetic dataset, we constructed 875 pairs covering 175 shapes from 23 categories. \rev{We refer readers to the supplementary material for a visualization of this test dataset, included a list of object categories in it. Note that our synthetic dataset has only one category overlapping with our training dataset (laptop). The rest of the categories are particularly useful for testing  the cross-category generalization ability of all the different techniques included in our evaluation.}

\vspace{-0.15cm}
\paragraph{Real Dataset.}  We also collected real data for evaluation under two different settings. 1) Real scan pairs of the same object but with different articulations(\textbf{RP2P}) 2) 3D pairs consisting of a full CAD model downloaded online, and a partial scan captured from the real world. In this case, the pair represents geometrically different objects under different articulations (\textbf{RF2P}). For \textbf{RP2P}, we collected 231 scan pairs from 10 categories. We manually segmented the scans into their moving parts. For \textbf{RF2P} we collected 150 pairs from 10 categories and also manually segmented the CAD models into moving parts. The testing categories in all our test datasets are different from the ones used for training. We also note that in this setting, there is a potential domain shift in our testing since our training data do not necessarily include realistic motion, as in the case of real data.


\vspace{-0.15cm}
\subsection{Deformation Flow}
We first evaluate our predicted deformation flow on two synthetic datasets \textbf{SF2F} and \textbf{SF2P} (where ground-truth flow is available). 

\vspace{-0.15cm}
\paragraph{Methods.} We test our method against various alternatives, including both learning and non-learning approaches. Specifically, we compare with three learning approaches including a 3D scene flow estimation approach (3DFlow) \cite{liu20183dflow}, feature-based matching with learned descriptors from a volumetric CNN (3DMatch) \cite{zeng20163dmatch} and from a multi-view CNN (LMVCNN) \cite{Huang:2017:LMVCNN}. For all three approaches, we train the corresponding networks from scratch using the same training data as ours. \rev{For 3DMatch and LMVCNN, we extracted point-wise descriptors with the learned networks and estimated the flow by finding nearest neighboring points in descriptor space}. In addition, we also compare with two non-learning approaches \cite{Sumner:2007} and \cite{huang2008non}, which leveraged non-rigid deformation for deformation flow estimation. We will refer to \cite{Sumner:2007} as ED and refer to \cite{huang2008non} as NRR in our baseline comparison. 

\vspace{-0.15cm}
\paragraph{Metrics.} We use two popular measures to evaluate the flow predicted by all methods. First, we use \textit{End-Point-Error (EPE)} as defined in \cite{yan2016scene}, which has been widely used for optical flow and scene flow evaluation. To be specific, given a ground truth flow field $\bff^{(gt)}$ and a predicted flow field $\bff$, the EPE is $E_{EPE}=\frac{1}{n}\overset{n}{\underset{i=1}{\sum}}\sqrt{(\bff_i-\bff_i^{gt})^2}$. We also use  the \textit{Percentage of Correct Correspondences (PCC)} curve as described in \cite{kim2011blended}, where the percentage of correspondences that are consistent with ground truth under different prescribed distances is shown. 

\begin{table}
\centering
\small
\begin{tabular}{|c|c|c|c|c|c|c|}
\hline
 & Ours &  3DFlow & ED & NRR & 3DMatch & LMVCNN \\ \hline
\textbf{SF2F} & \textbf{0.0210} &  0.0536 & 0.0481 & 0.0394 & 0.0715 & 0.0582 \\
\textbf{SF2P} & \textbf{0.0422} & 0.0892 & 0.0805 & 0.0556 & 0.138 & 0.093 \\\hline
\end{tabular}
\caption{EPE evaluation on dataset SF2F and SF2P for  all competing methods. EPE measures the Euclidean distance between predicted flow and the ground-truth flow. Smaller EPE means more accurate flow prediction. }
\label{tab:flow_eva_baseline}
\vspace{-0.65cm}
\end{table}


\begin{figure}[b!]
    \centering
    \includegraphics[width=\linewidth]{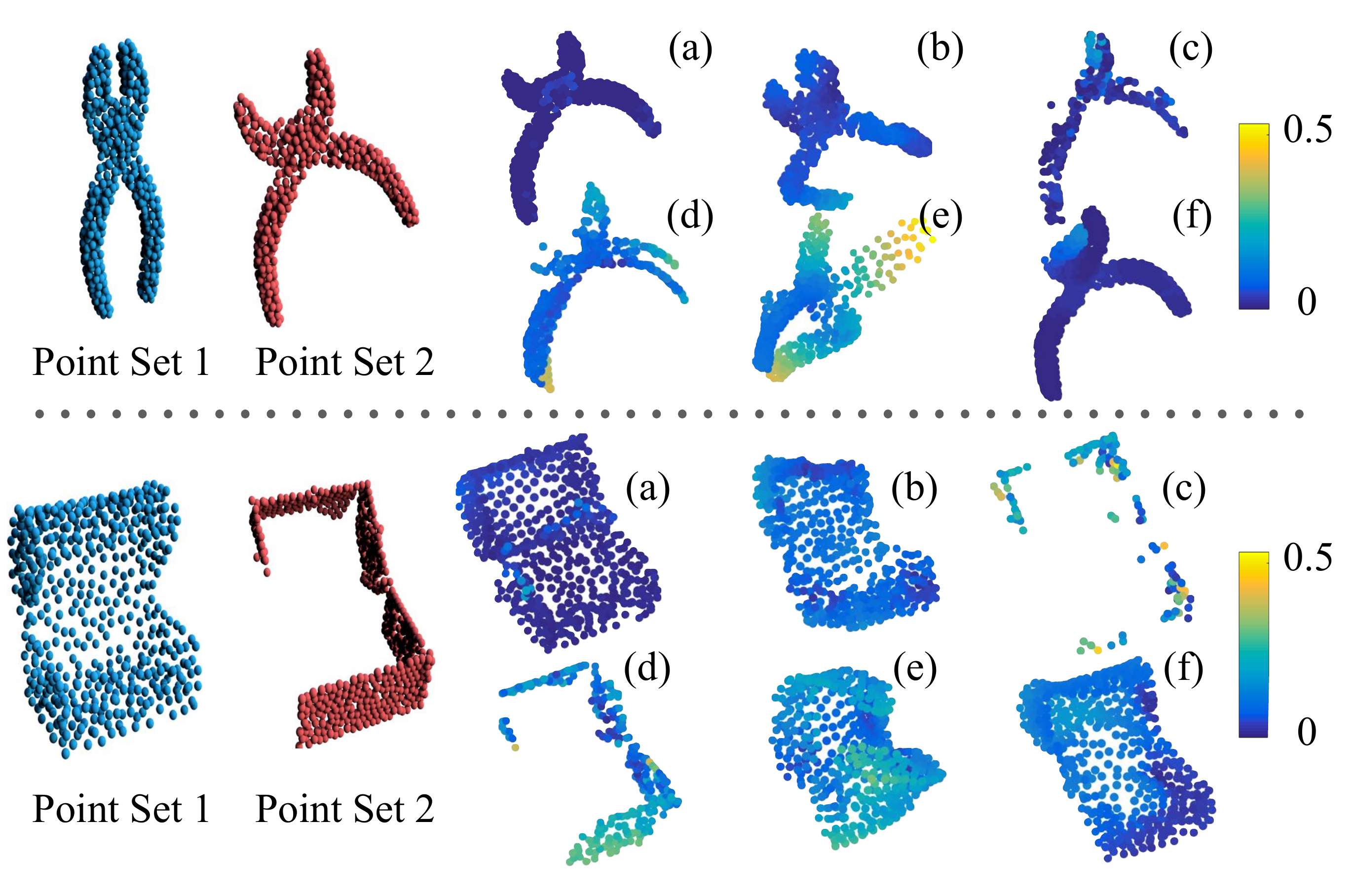}
    \vspace{-0.65cm}
    \caption{Deformation flow visualization. We estimate a dense flow from the point set 1 to point set 2 and apply the flow to deform the point set 1. Deformation results are shown from (a) to (f): (a) Ours, (b) 3DFlow, (c) 3D match, (d) LMVCNN, (e) ED, (f) NRR. Colors on the deformed point set denote the flow error of each point, whose range is shown on the color bar at the right side of each row.}
    \label{fig:corrs_vis}
    \vspace{-0.45cm}
\end{figure}

\vspace{-0.15cm}
\paragraph{Results.} We show the comparison of different approaches in Table~\ref{tab:flow_eva_baseline} in the case of the EPE metric. Our approach outperforms all the baseline methods by a large margin. \textbf{SF2F} contains pairs of full shapes from the same objects with different articulations while pairs in \textbf{SF2P} include one full shape and one partial scan. Due to the large missing data, the flow predictions of all approaches are less accurate on \textbf{SF2P} compared with those on \textbf{SF2F}, but still, our approach demonstrates more robustness and achieves the best performance. We visualize the deformation flow predicted by various approaches in Figure~\ref{fig:corrs_vis}. Similar to us, 3DFlow leverages PointNet in their scene flow estimation scheme. Their approach tends to have degraded performance while dealing with large motion, especially large rotations, which is a common scenario in part motions for man-made objects. They do not leverage motion structure between two frames, such as piecewise rigidity. In addition, they do not model missing data from one point cloud to another, resulting in a much worse performance on \textbf{SF2P} compared to our approach. 3DMatch and LMVCNN are not very suitable for dense flow estimation as they suffer from ambiguities due to symmetry, cannot handle  missing data well and lack smoothness in their flow prediction. Deformation-based approaches (ED, NRR) ignore the piece-wise rigidity property of objects and also result in artifacts when the articulation difference in the input pair is large, where a good set of initial correspondences becomes hard to estimate. We also note that given large portions of missing data, deformation-based approaches tend to generate largely unrealistic deformations around missing shape regions. Compared to all the above methods, our approach is able to parse the piecewise rigid motion of the objects much more reliably. 
Due to the explicit modeling of missing data, our approach avoids deforming the portion of point set $1$ which has no correspondences in point set $2$, and instead tends to generate the flow field via considering the other points on the same rigid part.

\begin{figure}[t!]
    \centering
    \includegraphics[width=\linewidth]{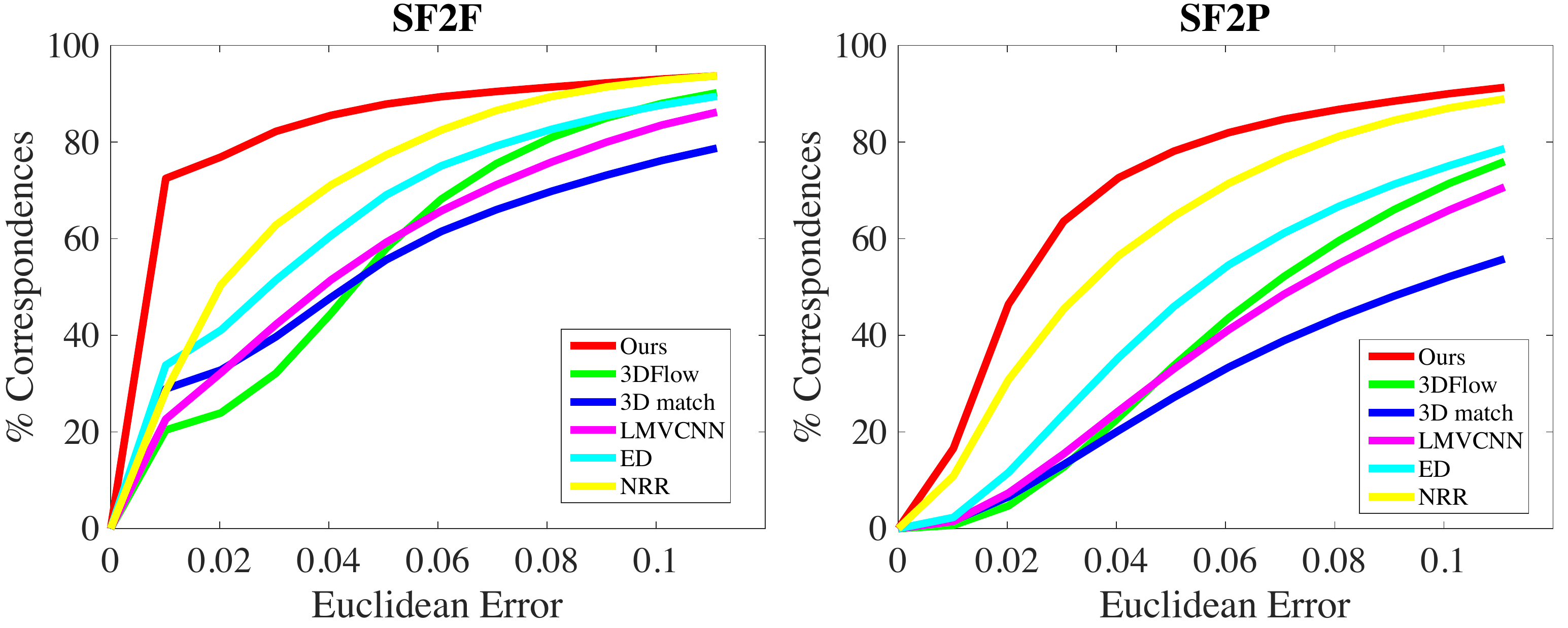}
    \vspace{-0.55cm}
    \caption{Percent of correspondences which have Euclidean error smaller than a threshold. The x-axis corresponds to different thresholds.}
    \label{fig:pcc}
    \vspace{-0.25cm}
\end{figure}

Figure~\ref{fig:pcc} shows the PCC curves of different approaches. We again observe that our method outputs more accurate flow estimation both in the local and global sense compared to other approaches.

\vspace{-0.15cm}
\subsection{Segmentation}
We then evaluate our segmentation performance both on synthetic datasets \textbf{SF2F} and \textbf{SF2P} and real datasets \textbf{RF2F} and \textbf{RF2P}. 

\vspace{-0.15cm}
\paragraph{Methods.} To ensure a fair comparison, we compare our approach with several other co-segmentation/motion segmentation baselines using our deformation flow prediction. Since our segmentation module can be regarded as a neural network-based version of sequential RANSAC with learnable hypothesis generation, verification and selection steps, we implemented a sequential RANSAC baseline, where we repeat the following steps until a stopping criterion is met: (a) figure out the largest rigid motion mode as well as its support in the current point set from the deformation flow; (b) remove all the supporting points from the discovered mode. We set the stopping criterion to be either a maximum number  of iterations has been met (10 in our case), or the remaining points are less than $5\%$ of the initial point set. Once we discovered the dominant motion modes as well as their associated supporting points, we assign labels to the rest of the points according to their closest motion modes. In addition, we also implemented several other baselines for comparison, including a spectral clustering approach (SC) \cite{tzionas2016reconstructing} and a JLinkage clustering approach (JLC)\cite{yuan2016space}. The spectral clustering approach leverages the fact that two points belonging to the same rigid part should maintain their Euclidean distance as well as the angular distance between their normals before and after deformation.
The JLinkage clustering approach samples a large number of motion hypotheses first and associates each data point with a hypotheses set. The closeness among data points can be defined based on the hypotheses sets and an iterative merging step is adopted to generate the final segmentation.
We also compare with the simultaneous flow estimation and segmentation approach (NRR) by Huang et al. \shortcite{huang2008non}. We found their segmentation results performs better using their own flow prediction (yet, the segmentation results are much worse than ours in any case). Therefore we report their segmentation results based on their own flow prediction. 
\begin{figure*}
    \centering
    \includegraphics[width=0.9\linewidth]{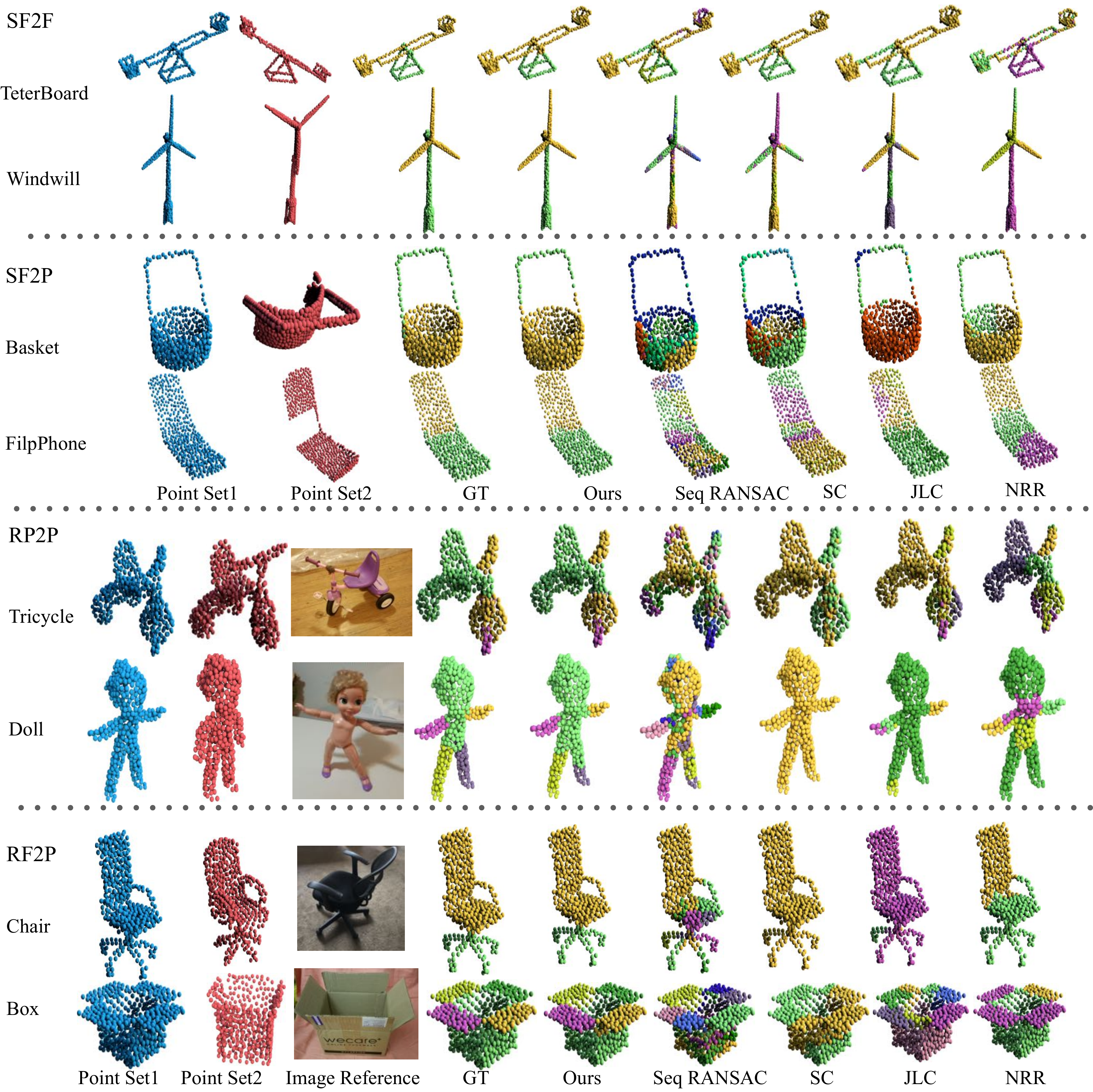}
    \vspace{-0.25cm}
    \caption{Segmentation visualization for all competing methods on synthetic and real data.}
    \label{fig:seg_vis}
    \vspace{-0.4cm}
\end{figure*}

\vspace{-0.15cm}
\paragraph{Metrics.} We use two evaluation metrics: Rand Index (RI) used in \cite{chen2009benchmark} and average per-part intersection over union (IoU) used in \cite{yi2016scalable}. Rand index is a similarity measurement between two data clusterings. We use the implementation provided by \cite{chen2009benchmark}. Average per-part IoU is a more sensitive metric to small parts. To compute per-part IoU between a set of ground truth segments $\{\hat{\by}_k\}_{k=1...K}$ and the predicted segments $\{\by_t\}_{t=1...T}$, we first use the Hungarian algorithm to find a bipartite matching between the ground-truth segment indicator vectors $\{\hat{\by}_k\}$ and the predicted segment indicator vectors $\{\by_t\}$ so that $\by_{t(k)}$ denotes the match of $\hat{\by}_k$. We then compute the per-part IoU as: $IoU(\by_t(k), \hat{\by}_{k})=\frac{<\by_t(k), \hat{\by}_{k}>}{||\by_t(k)||_1+|| \hat{\by}_{k}||_1-<\by_t(k),\hat{\by}_{k}>}$. If a part $\hat{\by}_k$ in the ground truth set has no match in the prediction set (the number of predicted parts is less than the ground truth), we  count its $IoU(\by_t(k), \hat{\by}_{k})$ as $0$. The final average per-part IoU is simply an average of the above IoU for all parts and all shapes.
\begin{table}[t!]
\centering
\small
\begin{tabular}{|c|c|c|c|c|c|}
\hline
 & SeqRANSAC & SC & JLC & NRR & Ours\\ \hline
SF2F & 60.2/55.8 & 80.6/69.4 & 74.7/67.3 & 74.1/57.3 & \textbf{83.8}/\textbf{77.3} \\
SF2P & 48.2/37.6 & 67.0/55.6 & 66.2/58.2 & 72.7/53.9 & \textbf{75.6}/\textbf{66.6} \\
RS2S & 56.7/43.0 & 79.1/67.1 & 80.4/73.0 & 78.4/65.5 & \textbf{88.3}/\textbf{83.5} \\
RF2S & 58.7/44.2 & 71.9/53.6 & 72.7/58.4 & 72.8/54.7 & \textbf{87.6}/\textbf{81.8} \\\hline
\end{tabular}
\caption{RI and IoU evalution on both synthetic and real datasets for all competing methods. Numbers in each cell represent RI/IoU. Both RI and IoU measure the segmentation consistency. Higher RI and higher IoU mean better segmentation prediction.  }
\label{tab:seg_eva_ri_iou}
\vspace{-0.7cm}
\end{table}

\vspace{-0.15cm}
\paragraph{Results.} We compare our approach with various baseline methods on four different datasets including two synthetic ones (\textbf{SF2F},\textbf{SF2P}) and two real ones (\textbf{RP2P},\textbf{RF2P}). The results are reported in Table~\ref{tab:seg_eva_ri_iou}
using the rand index and average per-part IoU as the evaluation metrics. Our approach outperforms all the baseline methods by a large margin, both on synthetic and real datasets, especially when using the per-part IoU metric, which indicates our approach has a better ability to capture the correct number of parts and is more capable to detect small parts.

To better understand the performance gain of our approach, we visualize the prediction results of different methods on the two synthetic datasets in Figure~\ref{fig:seg_vis}. Since it is hard to acquire perfect flow field prediction, the segmentation approach needs to be robust to input noise and imperfect flow while generating the predictions. Sequential RANSAC can handle input noise to some degree by properly setting a noise threshold while generating inlier supports. However different input shape pairs seem to require different thresholds -- a single threshold fails to provide satisfactory results in all cases. Our implementation of sequential RANSAC uses a cross-validated threshold and it tends to generate small discontinuous pieces in the shown examples. The SC and JLC approaches predict segmentations mainly based on the motion cues with little consideration of the underlying geometry, therefore they can over-group parts, such as the platform with the fulcrum. Our segmentation module instead considers both the motion cue and the underlying geometry, thus it is able to generate the correct segmentation even if the input flow is noisy and imperfect. Moreover, the SC and JLC approaches both require hand-tuned thresholds, which are quite sensitive to different types of shapes. Again we set the thresholds via cross-validation and find that they usually fail to discover rigid parts. NRR  often leads to under-segmentations such as the USB example in Figure~\ref{fig:seg_vis} while in other cases, it results in over-segmentations such as the flip phone example.

The experiments on the two real datasets demonstrate that our approach, trained on synthetic data, is able to generalize to real scans. We also visualize prediction results on \textbf{RP2P} and \textbf{RF2P} in Figure~\ref{fig:seg_vis}. The \textbf{RP2P} dataset contains two scans of the same underlying object with different articulations. In the challenging tricycle example, our approach successfully segments out the two small pedals and groups them together through their motion pattern. None of the baseline methods are capable of achieving this. \rev{When the number of parts and DoFs  increase, such as the articulated doll example, the baseline approaches cannot generate a proper number of parts in contrast to ours}. The \textbf{RF2P} dataset contains pairs of shapes from different objects, which is very challenging since the flow field from the point set 1 to point set 2 contains both motion flow and geometric flow caused by their geometric difference. To downweigh the influence of geometric flows, we also tried to optimize our predicted deformation flow with an as-rigid-as-possible (ARAP) objective ~\cite{Sorkine:2007:ASM} to preserve the local geometry before passing it to various segmentation approaches. Baseline methods either under-segment or over-segment the point sets and the segmentation boundaries are also very noisy. Our segmentation module again demonstrates robustness and is more capable of predicting the number of segments properly and generating cleaner motion boundaries.

\begin{table}[t!]
\centering
\small
\begin{tabular}{|c|c|c|c|c|}
\hline
 & (a) & (b) & (c) & (d)  \\ \hline
SF2F & \textbf{0.0377} & 0.0495 & 0.046 &      0.044  \\
SF2P & \textbf{0.0522} &       0.0771 & 0.764 & 0.0759  \\\hline
\end{tabular}
\caption{\rev{Ablation study of our network: we evaluate the deformation flow generated by different variations (a)-(d) of our framework with a single iteration}, as explained in Sec~\ref{sec:abl}, using EPE as the metric.}
\label{tab:flow_eva_our}
\vspace{-0.8cm}
\end{table}

\subsection{Ablation Study}
\label{sec:abl}
\begin{figure}
    \centering
    \includegraphics[width=\linewidth]{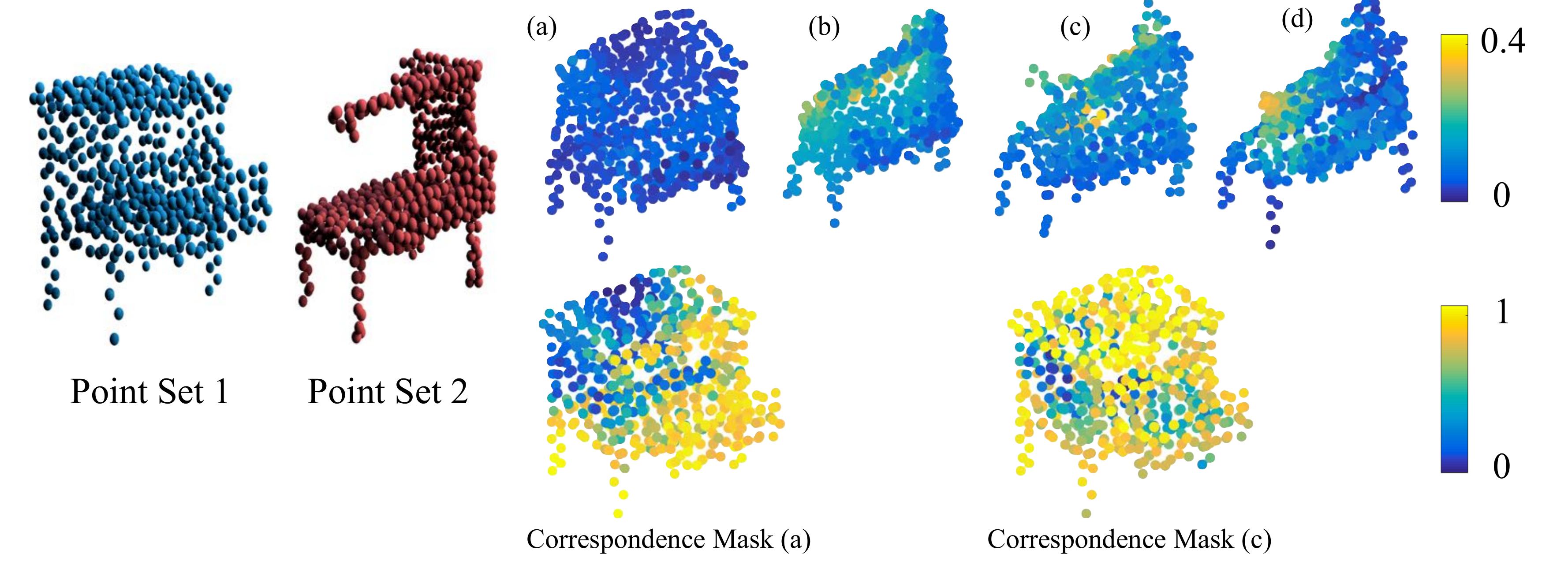}
    \caption{Deformation visualization for ablation study. We predict a deformation flow from the point set 1 to set 2 through different variations (a)-(d) of our framework, as explained in Sec~\ref{sec:abl}. Using the predicted flows to deform point set 1 generates results indexed from (a) to (d), where colors denote the deformation error. We also visualize the correspondence mask predicted by variation (a) and (c), where colors denote the soft mask value.}
    \label{fig:ablation_corrs_vis}
    \vspace{-0.25cm}
\end{figure}

We also conduct an ablation study by removing individual components of our framework. 
\rev{Specifically, we take a single iteration (one forward pass of the correspondence proposal module and flow module) and tested the deformation flow field generated by our framework using EPE: (a) with all the designed modules, (b) without correspondence mask predication, (c) without matching probability and correspondence mask supervision and  (d) without the flow module.}
In case (d), we simply compute the matching point for each point in set 1 using a weighted average of the points in set 2 based on the predicted matching probability. The results are presented in Table  ~\ref{tab:flow_eva_our}. We also visualize the predicted deformation flow from different settings in Figure~\ref{fig:ablation_corrs_vis}.

On both \textbf{SF2F} and \textbf{SF2P}, our full iterative pipeline achieves the best performance, which indicates the importance of leveraging segmentation to enforce piecewise rigidity on the predicted flow field. This also justifies our mutual reinforcement framework between correspondence estimation and part segmentation. Comparing (a) and (b), it can be seen that it is crucial to explicitly model the missing data aspect by introducing a correspondence mask prediction. This particularly improves the performance when matching a full shape and a partial shape from \textbf{SF2P}, but also improves matching quality between a full shape pair from \textbf{SF2F}. We suspect that this is because the network can decide which matching point pairs to trust in order to generate a deformation flow field. The predicted correspondence mask from (a) is visualized in Figure~\ref{fig:ablation_corrs_vis} and it successfully highlights regions on point set 1 where a matching point can be found on point set 2. Comparing (a) and (c), we see the importance of adding explicit supervision for matching probability and the correspondence mask prediction. From Figure~\ref{fig:ablation_corrs_vis} we see that without explicit supervision, the predicted correspondence mask is less faithful, resulting in a performance degradation.  Comparing (a) and (d), we see that using the correspondence proposal module results in less smooth deformations and it cannot handle missing areas well, which justifies the use of an additional flow module.

\begin{table}[t!]
\centering
\small
\begin{tabular}{|c|c|c|c|c|c|}
\hline
 & & 1 iter & 2iter & 5iter & 9iter \\ \hline
\multirow{ 3}{*}{SF2F} & EPE (Corrs) & 0.0377 & 0.0283 & 0.0222 & \textbf{0.0210}\\
& RI (Seg) & 0.741 & 0.783 & 0.809 & \textbf{0.838} \\
& IoU (Seg) & 0.631 & 0.697 & 0.737 & \textbf{0.773} \\\hline
\multirow{ 3}{*}{SF2P} & EPE (Corrs) & 0.0522 & 0.0464 & 0.0429 & \textbf{0.0422} \\
& RI (Seg) & 0.701 & 0.73 & 0.746 & \textbf{0.756} \\
& IoU (Seg) & 0.577 & 0.623 & 0.653 & \textbf{0.666} \\\hline
\end{tabular}
\caption{Iterative improvements of our method on two synthetic datasets, evaluated on both shape correspondences and segmentation. }
\label{tab:iterative_refine}
\vspace{-0.8cm}
\end{table}

\vspace{-0.25cm}
\rev{
\paragraph{Recurrent Unit Design in the Segmentation Module.}
Instead of  using our RPEN\ design that  incorporates explicit memory to encode the segmentation progress as explained in Section \ref{sec:segmodule}, an alternative approach would be to use LSTMs as recurrent units in the segmentation module. We use RI/IoU to evaluate the segmentation score. We found that using an LSTM in the segmentation module achieves a score of 0.740/0.584 on \textbf{SF2F}, while our design achieves a higher score of 0.838/0.773.
}

\vspace{-0.15cm}
\paragraph{Iterative Flow and Segmentation Estimation.}
\rev{To validate whether correspondence proposals, flow estimation and part segmentation operate in a mutually reinforcing way, we tested our iterative framework on two synthetic datasets \textbf{SF2F} and \textbf{SF2P}. We show  how flow and segmentation estimation improves with more iterations quantitatively in Table~\ref{tab:iterative_refine}. We also refer to Figure~\ref{fig:iter_vis} for qualitative results. The predicted flow tends to non-rigidly deform the object at early iterations and then becomes increasingly piece-wise rigid.}

\vspace{-0.15cm}
\subsection{Comparison with Other Learning-Based Motion Segmentation Approaches}
\begin{figure}[t!]
    \centering
    \includegraphics[width=\linewidth]{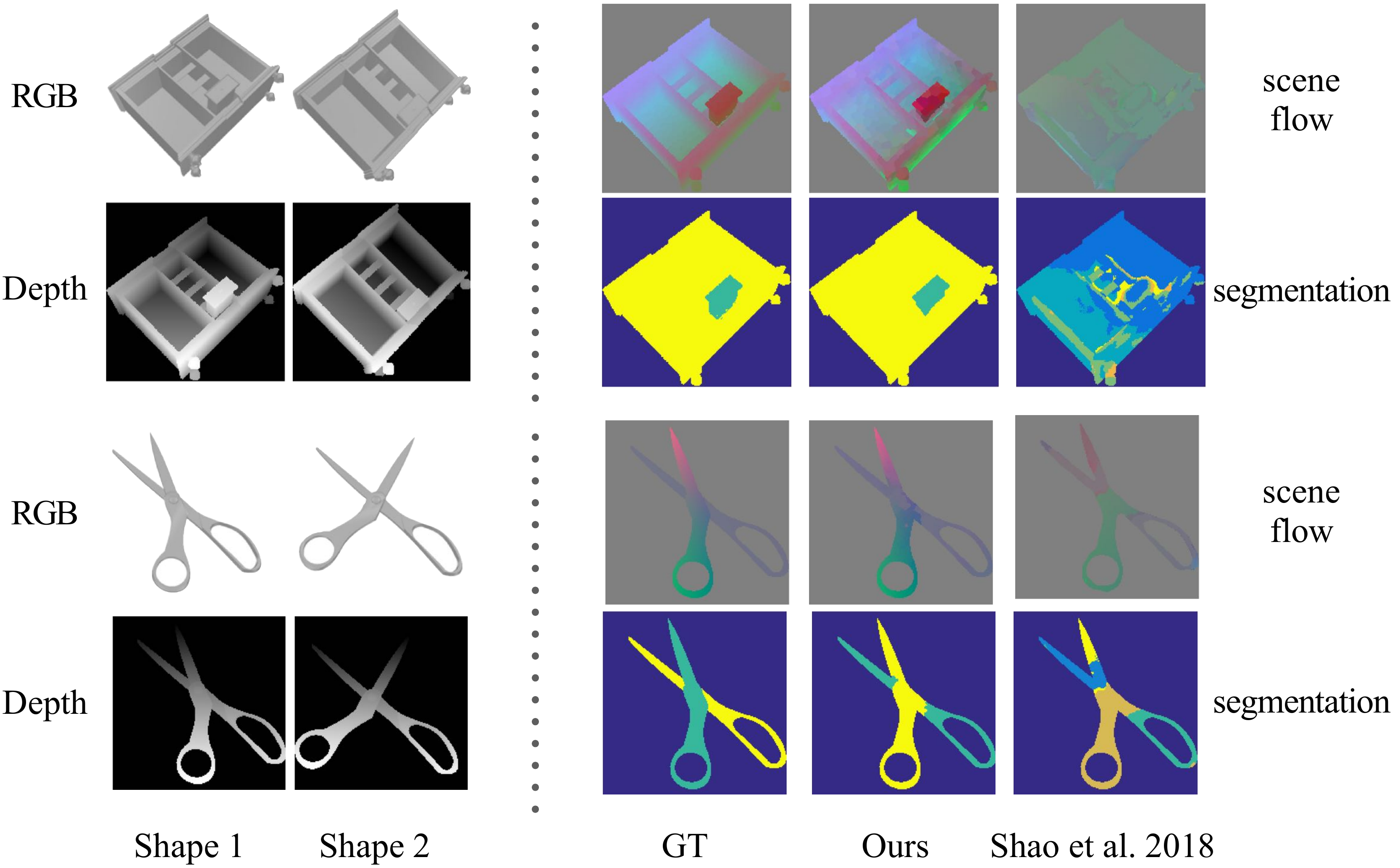}
    \vspace{-0.65cm}
    \caption{Comparison with \cite{shao2018motion}. The input includes a pair of RGB and a pair of depth images representing the underlying shape pair. Both \cite{shao2018motion} and our method predict a scene flow plus a motion based segmentation, visualized here together with the ground truth (GT).}
    \vspace{-0.25cm}
    \label{fig:shao_comp}
\end{figure}

We compare here our method with the concurrent learning method by Shao et al. \shortcite{shao2018motion}. Their method trains a joint flow estimation and segmentation network for motion-based object segmentation. They take an RGB-D pair as input, and they convert the depth image into a partial point cloud with known camera parameters. Then their network consumes the RGB information as well as the point clouds and generates a motion-based segmentation for the input frames. To compare with that approach, we use our synthetic dataset \textbf{SP2P}, where we can render CAD models with different articulations into RGBD pairs and apply both their approach and ours for motion segmentation. Note that our approach only uses the point cloud information, while their approach exploits both the rendered images as well as the partial point clouds. We also rendered our training data into RGB-D pairs, and asked the authors of \cite{shao2018motion} to re-train their network on our training set so that they can handle the CAD rendered images. The quantitative comparison is shown in Table~\ref{tab:learning_comp}. Our flow field and segmentation estimation outperforms \cite{shao2018motion} by a large margin. We visualize the prediction results in Figure~\ref{fig:shao_comp}. \cite{shao2018motion} cannot reliably estimate flows for complex structures, in particular in texture-less settings, such as the drawer in a cabinet. This leads to largely inaccurate segmentation results. Our approach instead fully operates on 3D and is more effective to capture the object structure. In addition, \cite{shao2018motion} cannot group object parts whose centers are very close to each other and do not move much, e.g. the scissors example.

\begin{table}[b!]
\centering
\small
\begin{tabular}{|c|c|c|c|}
\hline
SP2P & EPE (Corrs) & RI (Seg) & IoU (Seg) \\ \hline
\cite{shao2018motion} & 0.0862 & 0.686 & 0.563\\
Ours & \textbf{0.0369} & \textbf{0.833} & \textbf{0.756}\\
\hline
\end{tabular}
\caption{Numerical evaluation compared to \cite{shao2018motion}.}
\label{tab:learning_comp}
\end{table}

\subsection{Computational cost}
\rev{Since our network operates on representations capturing pairwise relationships between input shape points, the theoretical complexity of both training and testing stage is $O(N^2)$ (where $N$ is the number of input points). However, we found that our method is still quite fast: it takes 0.4 seconds to process a shape pair with a single correspondence proposal, flow estimation and segmentation pass on average, tested on a NVIDIA TITAN Xp GPU. We also noticed that increasing the number of points results in  small improvements  e.g., we found that using 1024 points increases the segmentation Rand Index  from $83.8\%$ to $84.2\%$ in the SF2F dataset.}

\vspace{-0.15cm}
\subsection{Applications}
Our framework co-analyzes a pair of shapes, generating a dense flow field from one point set to another and also the motion-based part segmentation of the two shapes.
These outputs essentially reveal the functional structure of the underlying dynamic objects and can benefit various applications we discuss below.

\begin{figure}[b!]
    \centering
    \includegraphics[width=\linewidth]{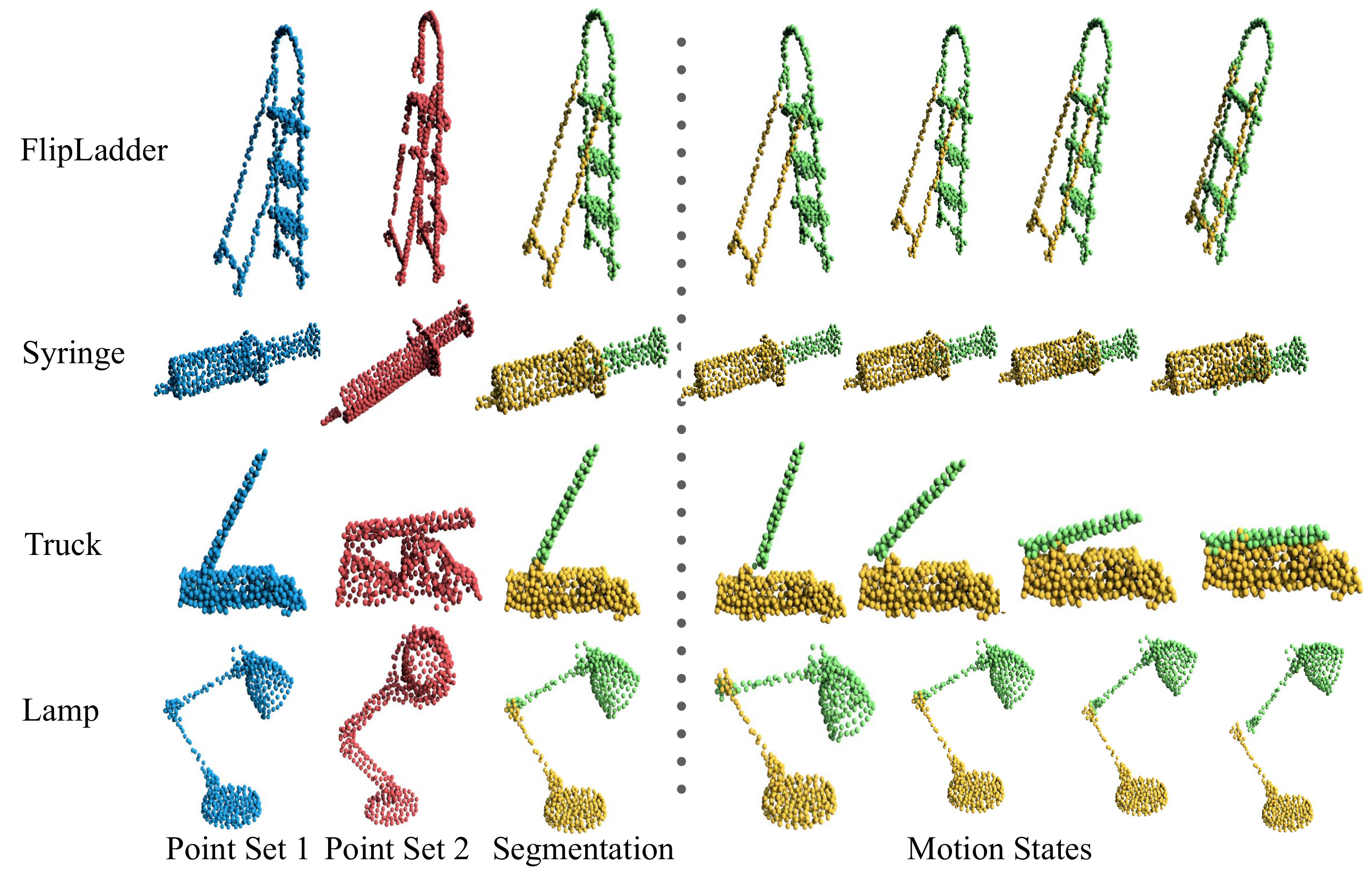}
    \caption{Our method can be used for shape animation. Given the predicted segmentation of articulated parts (left), one can generate a sequence of animated shapes by interpolating the motion (right).}
    \label{fig:app_animation}
    \vspace{-0.3cm}
\end{figure}

\vspace{-0.15cm}
\paragraph{Shape Animation}
The output of our framework can be directly used to animate shapes. Given a shape pair $(P,Q)$, we co-segment them into rigid parts $\{P_t\}$ and also estimate the corresponding rigid motions $\{H_t\}$
We can then interpolate between the motion states of $P$ and $Q$, generating in-between motion frames for $P$. To interpolate between two rigid motions $H_i$ and $H_j$, we sample the geodesic paths connecting $H_i$ and $H_j$ following \cite{vzefran1998interpolation}. Assuming 4x4 homogeneous
matrices representation for $H_i$ and $H_j$, an interpolation between $H_i$ and $H_j$ can be computed as $ F(t; H_i, H_j) = \text{exp}(t\text{log}(H_jH_i^{-1}))H_i,\;t \in [0, 1]$. We fix the motion for a selected part and generate different motion states for the rest of the parts, \rev{and visualize our results in Figure \ref{fig:app_animation}}. We are able to animate both revolute joints and prismatic joints. Such animation reveals the underlying functionality of the object and is useful for adding interactivity to that object in a virtual environment.


\begin{figure}[t!]
    \centering
    \includegraphics[width=\linewidth]{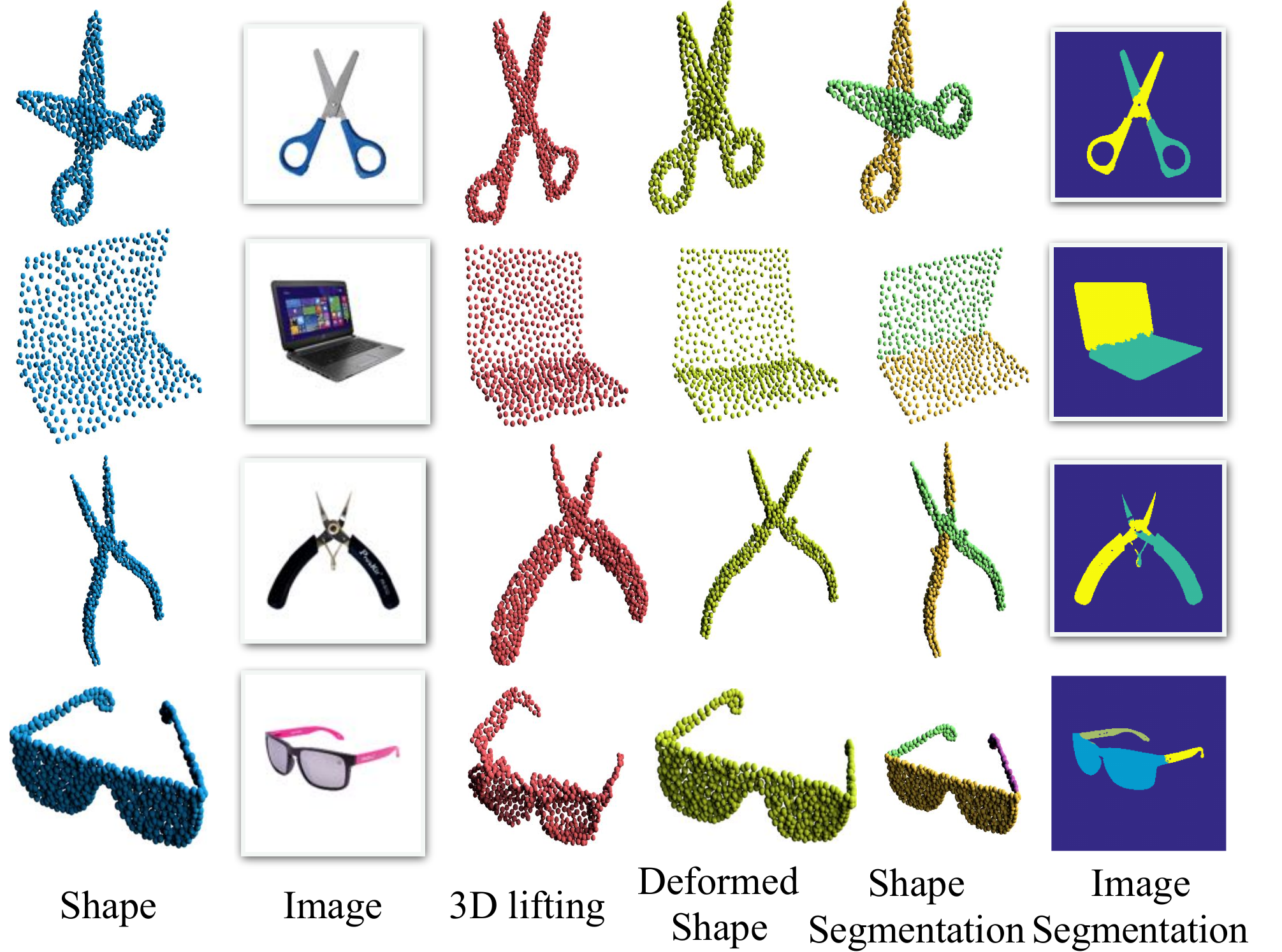}
    \caption{Our method can be used for joint shape-image analysis. Given a pair including a 3D shape and a 2D image, we can jointly align them and output the segmentation for both the shape and the image.}
    \label{fig:app_s2i}
\end{figure}

\vspace{-0.15cm}
\paragraph{Part Induction from a Shape-Image Pair}
Recently we have witnessed a lot of progress in single image-based 3D reconstruction, which opens up a new application of our framework, namely joint motion segmentation for a shape and image pair. Given an articulated 3D shape and a related 2D product image as input, we lift the 2D image onto 3D and then apply our framework to co-segment the lifted 3D shape and the input 3D shape. The segmentation information can be later propagated back from the lifted 3D shape back onto the 2D image, resulting in a motion segmentation for the image as well. This results in a co-segmentation for both the 3D shape and the 2D image based on discovered motion cues. This setting is attractive since given an articulated 3D shape of a man-made object, we can easily find lots of images online describing related objects with different articulation states through product search. Being able to extract motion information for these shape image pairs benefit a detailed functional understanding of both domains. For the purpose of lifting the 2D product image onto 3D, we design another neural network and we refer to the \emph{supplementary} for more details on this lifting network.

We visualize our shape-image pair co-analysis in Figure~\ref{fig:app_s2i}. In each row, we are given an input shape plus an input product image describing a related but different object. \rev{We first convert the 2D image into a 3D point set using the lifting network. The lifted 3D shapes are visualized in the $3$rd column. We then apply our framework for the input pair, resulting in a deformed version of the input shape visualized in the $4$th column. We note that articulation of the deformed input shape becomes  more similar to the one of the lifted point set. We visualize the segmentation for the input shape in the $5$th column and we propagate the segmentation of the lifted point set back to the 2D image (our lifting procedure maintains the correspondence between the image pixels and the lifted points), which is shown in the last column.}


\section{Limitations and Future Work}
\label{sec:limit}
\rev{We presented a neural network architecture that is able to discover parts of objects by analyzing the underlying  articulation states and geometry of different shapes}. Our network is able to generalize to novel objects and classes not observed during training. 

There are several avenues for future research directions. First, our method uses 3D point cloud representations, which might under-sample small parts, or cannot deal with rotationally symmetric parts, such as bottle caps. \rev{We also found that parts that slide towards the interior of a shape, such as sliding knives, are more challenging since point samples on folding shape layers are hard to differentiate
}.  Increasing the point cloud resolution or introducing an attention mechanism that results in a dynamic adaptation of resolution could help to deal with these cases. Second, part induction between different shape instances with large topology variation remains a challenging problem. The point-wise deformation flow itself becomes harder to define in these cases. When it comes to large geometric differences, the segmentation module would need to be redesigned to deal with both local rigid and non-rigid deformation flow. \rev{Finally, our method currently infers parts and motions from pairs of shapes. In the future, it would be interesting to infer those from a single input, or analyze larger sets of objects to discover common articulation patterns within a shape family.}
 

\section*{Acknowledgements}
\rev{This research was supported by NSF grants DMS-1521608,  IIS-1528025, IIS-1763268, CHS-1617333, the Stanford AI Lab-Toyota Center for Artificial Intelligence Research, as well as gifts from Adobe and Amazon AWS. We thank Yang Zhou, Zhan Xu, and Olga Vesselova for helping with the object scans.}


\bibliographystyle{ACM-Reference-Format}
\bibliography{part_mobility}


\begin{thebibliography}{68}


\ifx \showCODEN    \undefined \def \showCODEN     #1{\unskip}     \fi
\ifx \showDOI      \undefined \def \showDOI       #1{#1}\fi
\ifx \showISBNx    \undefined \def \showISBNx     #1{\unskip}     \fi
\ifx \showISBNxiii \undefined \def \showISBNxiii  #1{\unskip}     \fi
\ifx \showISSN     \undefined \def \showISSN      #1{\unskip}     \fi
\ifx \showLCCN     \undefined \def \showLCCN      #1{\unskip}     \fi
\ifx \shownote     \undefined \def \shownote      #1{#1}          \fi
\ifx \showarticletitle \undefined \def \showarticletitle #1{#1}   \fi
\ifx \showURL      \undefined \def \showURL       {\relax}        \fi
\providecommand\bibfield[2]{#2}
\providecommand\bibinfo[2]{#2}
\providecommand\natexlab[1]{#1}
\providecommand\showeprint[2][]{arXiv:#2}

\bibitem[\protect\citeauthoryear{Bogo, Kanazawa, Lassner, Gehler, Romero, and
  Black}{Bogo et~al\mbox{.}}{2016}]%
        {Bogo:ECCV:2016}
\bibfield{author}{\bibinfo{person}{Federica Bogo}, \bibinfo{person}{Angjoo
  Kanazawa}, \bibinfo{person}{Christoph Lassner}, \bibinfo{person}{Peter
  Gehler}, \bibinfo{person}{Javier Romero}, {and} \bibinfo{person}{Michael~J.
  Black}.} \bibinfo{year}{2016}\natexlab{}.
\newblock \showarticletitle{Keep it {SMPL}: Automatic Estimation of {3D} Human
  Pose and Shape from a Single Image}. In \bibinfo{booktitle}{\emph{Proc.
  ECCV}}.
\newblock


\bibitem[\protect\citeauthoryear{Boscaini, Masci, Melzi, Bronstein, Castellani,
  and Vandergheynst}{Boscaini et~al\mbox{.}}{2015}]%
        {Boscaini2015spectral}
\bibfield{author}{\bibinfo{person}{D. Boscaini}, \bibinfo{person}{J. Masci},
  \bibinfo{person}{S. Melzi}, \bibinfo{person}{M.~M. Bronstein},
  \bibinfo{person}{U. Castellani}, {and} \bibinfo{person}{P. Vandergheynst}.}
  \bibinfo{year}{2015}\natexlab{}.
\newblock \showarticletitle{Learning Class-specific Descriptors for Deformable
  Shapes Using Localized Spectral Convolutional Networks}. In
  \bibinfo{booktitle}{\emph{Proc. SGP}}.
\newblock


\bibitem[\protect\citeauthoryear{Boykov, Veksler, and Zabih}{Boykov
  et~al\mbox{.}}{2001}]%
        {Boykov2001}
\bibfield{author}{\bibinfo{person}{Yuri Boykov}, \bibinfo{person}{Olga
  Veksler}, {and} \bibinfo{person}{Ramin Zabih}.}
  \bibinfo{year}{2001}\natexlab{}.
\newblock \showarticletitle{Efficient Approximate Energy Minimization via Graph
  Cuts}.
\newblock \bibinfo{journal}{\emph{IEEE Transactions on Pattern Analysis and
  Machine Intelligence}} \bibinfo{volume}{20}, \bibinfo{number}{12}
  (\bibinfo{year}{2001}), \bibinfo{pages}{1222--1239}.
\newblock


\bibitem[\protect\citeauthoryear{Brachmann, Krull, Nowozin, Shotton, Michel,
  Gumhold, and Rother}{Brachmann et~al\mbox{.}}{2017}]%
        {brachmann2017dsac}
\bibfield{author}{\bibinfo{person}{Eric Brachmann}, \bibinfo{person}{Alexander
  Krull}, \bibinfo{person}{Sebastian Nowozin}, \bibinfo{person}{Jamie Shotton},
  \bibinfo{person}{Frank Michel}, \bibinfo{person}{Stefan Gumhold}, {and}
  \bibinfo{person}{Carsten Rother}.} \bibinfo{year}{2017}\natexlab{}.
\newblock \showarticletitle{DSAC-Differentiable RANSAC for camera
  localization}. In \bibinfo{booktitle}{\emph{Proc. CVPR}}.
\newblock


\bibitem[\protect\citeauthoryear{Chang, Funkhouser, Guibas, Hanrahan, Huang,
  Li, Savarese, Savva, Song, Su, et~al\mbox{.}}{Chang et~al\mbox{.}}{2015}]%
        {chang2015shapenet}
\bibfield{author}{\bibinfo{person}{Angel~X Chang}, \bibinfo{person}{Thomas
  Funkhouser}, \bibinfo{person}{Leonidas Guibas}, \bibinfo{person}{Pat
  Hanrahan}, \bibinfo{person}{Qixing Huang}, \bibinfo{person}{Zimo Li},
  \bibinfo{person}{Silvio Savarese}, \bibinfo{person}{Manolis Savva},
  \bibinfo{person}{Shuran Song}, \bibinfo{person}{Hao Su}, {et~al\mbox{.}}}
  \bibinfo{year}{2015}\natexlab{}.
\newblock \showarticletitle{Shapenet: An information-rich 3d model repository}.
\newblock \bibinfo{journal}{\emph{arXiv preprint arXiv:1512.03012}}
  (\bibinfo{year}{2015}).
\newblock


\bibitem[\protect\citeauthoryear{Chang, Li, Mitra, Pauly, and Wand}{Chang
  et~al\mbox{.}}{2012}]%
        {dgp:2012:star}
\bibfield{author}{\bibinfo{person}{Will Chang}, \bibinfo{person}{Hao Li},
  \bibinfo{person}{Niloy Mitra}, \bibinfo{person}{Mark Pauly}, {and}
  \bibinfo{person}{Michael Wand}.} \bibinfo{year}{2012}\natexlab{}.
\newblock \showarticletitle{{Dynamic Geometry Processing}}. In
  \bibinfo{booktitle}{\emph{Eurographics 2012 - Tutorials}}.
\newblock


\bibitem[\protect\citeauthoryear{Chen, Golovinskiy, and Funkhouser}{Chen
  et~al\mbox{.}}{2009}]%
        {chen2009benchmark}
\bibfield{author}{\bibinfo{person}{Xiaobai Chen}, \bibinfo{person}{Aleksey
  Golovinskiy}, {and} \bibinfo{person}{Thomas Funkhouser}.}
  \bibinfo{year}{2009}\natexlab{}.
\newblock \showarticletitle{A benchmark for 3D mesh segmentation}.
\newblock  \bibinfo{volume}{28}, \bibinfo{number}{3} (\bibinfo{year}{2009}),
  \bibinfo{pages}{73}.
\newblock


\bibitem[\protect\citeauthoryear{Christoph, Schindler, and Stefan}{Christoph
  et~al\mbox{.}}{2015}]%
        {vogel2015sfe}
\bibfield{author}{\bibinfo{person}{Vogel Christoph}, \bibinfo{person}{Konrad
  Schindler}, {and} \bibinfo{person}{Roth Stefan}.}
  \bibinfo{year}{2015}\natexlab{}.
\newblock \showarticletitle{3D Scene Flow Estimation with a Piecewise Rigid
  Scene Model}.
\newblock \bibinfo{journal}{\emph{International Journal of Computer Vision}}
  \bibinfo{volume}{115}, \bibinfo{number}{1} (\bibinfo{year}{2015}).
\newblock


\bibitem[\protect\citeauthoryear{Fischler and Bolles}{Fischler and
  Bolles}{1981}]%
        {Fischler:1981:RSC}
\bibfield{author}{\bibinfo{person}{Martin~A. Fischler} {and}
  \bibinfo{person}{Robert~C. Bolles}.} \bibinfo{year}{1981}\natexlab{}.
\newblock \showarticletitle{Random Sample Consensus: A Paradigm for Model
  Fitting with Applications to Image Analysis and Automated Cartography}.
\newblock \bibinfo{journal}{\emph{Commun. ACM}} \bibinfo{volume}{24},
  \bibinfo{number}{6} (\bibinfo{year}{1981}).
\newblock


\bibitem[\protect\citeauthoryear{Golovinskiy and Funkhouser}{Golovinskiy and
  Funkhouser}{2009}]%
        {Golovinskiy:2009:CSM}
\bibfield{author}{\bibinfo{person}{Aleksey Golovinskiy} {and}
  \bibinfo{person}{Thomas Funkhouser}.} \bibinfo{year}{2009}\natexlab{}.
\newblock \showarticletitle{{C}onsistent {S}egmentation of {3D} {M}odels}.
\newblock \bibinfo{journal}{\emph{Computers \& Graphics}} \bibinfo{volume}{33},
  \bibinfo{number}{3} (\bibinfo{year}{2009}).
\newblock


\bibitem[\protect\citeauthoryear{Golyani, Kim, Maier, Nie{\ss}ner, Stricker,
  and Kautz}{Golyani et~al\mbox{.}}{2017}]%
        {golyanik2017multiframe}
\bibfield{author}{\bibinfo{person}{Vladislav Golyani}, \bibinfo{person}{Kihwan
  Kim}, \bibinfo{person}{Robert Maier}, \bibinfo{person}{Matthias Nie{\ss}ner},
  \bibinfo{person}{Didier Stricker}, {and} \bibinfo{person}{Jan Kautz}.}
  \bibinfo{year}{2017}\natexlab{}.
\newblock \showarticletitle{Multiframe Scene Flow with Piecewise Rigid Motion}.
  In \bibinfo{booktitle}{\emph{Proc. 3DV}}.
\newblock


\bibitem[\protect\citeauthoryear{Hassner, Harel, Paz, and Enbar}{Hassner
  et~al\mbox{.}}{2015}]%
        {hassner2015effective}
\bibfield{author}{\bibinfo{person}{Tal Hassner}, \bibinfo{person}{Shai Harel},
  \bibinfo{person}{Eran Paz}, {and} \bibinfo{person}{Roee Enbar}.}
  \bibinfo{year}{2015}\natexlab{}.
\newblock \showarticletitle{Effective face frontalization in unconstrained
  images}. In \bibinfo{booktitle}{\emph{Proc. CVPR}}.
\newblock


\bibitem[\protect\citeauthoryear{Hornácek, Fitzgibbon, and Rother}{Hornácek
  et~al\mbox{.}}{2014}]%
        {Hornacek2014}
\bibfield{author}{\bibinfo{person}{M. Hornácek}, \bibinfo{person}{A.
  Fitzgibbon}, {and} \bibinfo{person}{C. Rother}.}
  \bibinfo{year}{2014}\natexlab{}.
\newblock \showarticletitle{SphereFlow: 6 DoF Scene Flow from RGB-D Pairs}. In
  \bibinfo{booktitle}{\emph{Proc. CVPR}}.
\newblock


\bibitem[\protect\citeauthoryear{Hu, Fan, and Liu}{Hu et~al\mbox{.}}{2012}]%
        {Hu:2012:CSS}
\bibfield{author}{\bibinfo{person}{Ruizhen Hu}, \bibinfo{person}{Lubin Fan},
  {and} \bibinfo{person}{Ligang Liu}.} \bibinfo{year}{2012}\natexlab{}.
\newblock \showarticletitle{Co-Segmentation of 3D Shapes via Subspace
  Clustering}.
\newblock \bibinfo{journal}{\emph{Computer Graphics Forum}}
  \bibinfo{volume}{31}, \bibinfo{number}{5} (\bibinfo{year}{2012}).
\newblock


\bibitem[\protect\citeauthoryear{Hu, Li, Van~Kaick, Shamir, Zhang, and
  Huang}{Hu et~al\mbox{.}}{2017}]%
        {hu2017learning}
\bibfield{author}{\bibinfo{person}{Ruizhen Hu}, \bibinfo{person}{Wenchao Li},
  \bibinfo{person}{Oliver Van~Kaick}, \bibinfo{person}{Ariel Shamir},
  \bibinfo{person}{Hao Zhang}, {and} \bibinfo{person}{Hui Huang}.}
  \bibinfo{year}{2017}\natexlab{}.
\newblock \showarticletitle{Learning to predict part mobility from a single
  static snapshot}.
\newblock \bibinfo{journal}{\emph{ACM Transactions on Graphics}}
  \bibinfo{volume}{36}, \bibinfo{number}{6} (\bibinfo{year}{2017}),
  \bibinfo{pages}{227}.
\newblock


\bibitem[\protect\citeauthoryear{Huang, Kalogerakis, Chaudhuri, Ceylan, Kim,
  and Yumer}{Huang et~al\mbox{.}}{2017}]%
        {Huang:2017:LMVCNN}
\bibfield{author}{\bibinfo{person}{Haibin Huang}, \bibinfo{person}{Evangelos
  Kalogerakis}, \bibinfo{person}{Siddhartha Chaudhuri}, \bibinfo{person}{Duygu
  Ceylan}, \bibinfo{person}{Vladimir~G. Kim}, {and} \bibinfo{person}{Ersin
  Yumer}.} \bibinfo{year}{2017}\natexlab{}.
\newblock \showarticletitle{Learning Local Shape Descriptors from Part
  Correspondences with Multiview Convolutional Networks}.
\newblock \bibinfo{journal}{\emph{ACM Transactions on Graphics}}
  \bibinfo{volume}{37}, \bibinfo{number}{1} (\bibinfo{year}{2017}).
\newblock


\bibitem[\protect\citeauthoryear{Huang, Wang, and Guibas}{Huang
  et~al\mbox{.}}{2014}]%
        {Huang:2014:FMN}
\bibfield{author}{\bibinfo{person}{Qixing Huang}, \bibinfo{person}{Fan Wang},
  {and} \bibinfo{person}{Leonidas Guibas}.} \bibinfo{year}{2014}\natexlab{}.
\newblock \showarticletitle{Functional Map Networks for Analyzing and Exploring
  Large Shape Collections}.
\newblock \bibinfo{journal}{\emph{ACM Transactions on Graphics}}
  \bibinfo{volume}{33}, \bibinfo{number}{4} (\bibinfo{year}{2014}).
\newblock


\bibitem[\protect\citeauthoryear{Huang, Adams, Wicke, and Guibas}{Huang
  et~al\mbox{.}}{2008}]%
        {huang2008non}
\bibfield{author}{\bibinfo{person}{Qi-Xing Huang}, \bibinfo{person}{Bart
  Adams}, \bibinfo{person}{Martin Wicke}, {and} \bibinfo{person}{Leonidas~J
  Guibas}.} \bibinfo{year}{2008}\natexlab{}.
\newblock \showarticletitle{Non-rigid registration under isometric
  deformations}.
\newblock  \bibinfo{volume}{27}, \bibinfo{number}{5} (\bibinfo{year}{2008}),
  \bibinfo{pages}{1449--1457}.
\newblock


\bibitem[\protect\citeauthoryear{Jaimez, Souiai, Stueckler, Gonzalez-Jimenez,
  and Cremers}{Jaimez et~al\mbox{.}}{2015}]%
        {jaimez15_mocoop}
\bibfield{author}{\bibinfo{person}{M. Jaimez}, \bibinfo{person}{M. Souiai},
  \bibinfo{person}{J. Stueckler}, \bibinfo{person}{J. Gonzalez-Jimenez}, {and}
  \bibinfo{person}{D. Cremers}.} \bibinfo{year}{2015}\natexlab{}.
\newblock \showarticletitle{Motion Cooperation: Smooth Piece-Wise Rigid Scene
  Flow from RGB-D Images}. In \bibinfo{booktitle}{\emph{Proc. 3DV}}.
\newblock


\bibitem[\protect\citeauthoryear{James and Twigg}{James and Twigg}{2005}]%
        {James:2005:SMA}
\bibfield{author}{\bibinfo{person}{Doug~L. James} {and}
  \bibinfo{person}{Christopher~D. Twigg}.} \bibinfo{year}{2005}\natexlab{}.
\newblock \showarticletitle{Skinning Mesh Animations}.
\newblock \bibinfo{journal}{\emph{ACM Transactions on Graphics}}
  \bibinfo{volume}{24}, \bibinfo{number}{3} (\bibinfo{year}{2005}).
\newblock


\bibitem[\protect\citeauthoryear{Kalogerakis, Averkiou, Maji, and
  Chaudhuri}{Kalogerakis et~al\mbox{.}}{2017}]%
        {Kalogerakis:2017:ShapePFCN}
\bibfield{author}{\bibinfo{person}{Evangelos Kalogerakis},
  \bibinfo{person}{Melinos Averkiou}, \bibinfo{person}{Subhransu Maji}, {and}
  \bibinfo{person}{Siddhartha Chaudhuri}.} \bibinfo{year}{2017}\natexlab{}.
\newblock \showarticletitle{3{D} Shape Segmentation with Projective
  Convolutional Networks}. In \bibinfo{booktitle}{\emph{Proc. CVPR}}.
\newblock


\bibitem[\protect\citeauthoryear{Kim, Li, Mitra, Chaudhuri, DiVerdi, and
  Funkhouser}{Kim et~al\mbox{.}}{2013}]%
        {Kim:2013:LPT}
\bibfield{author}{\bibinfo{person}{Vladimir~G. Kim}, \bibinfo{person}{Wilmot
  Li}, \bibinfo{person}{Niloy~J. Mitra}, \bibinfo{person}{Siddhartha
  Chaudhuri}, \bibinfo{person}{Stephen DiVerdi}, {and} \bibinfo{person}{Thomas
  Funkhouser}.} \bibinfo{year}{2013}\natexlab{}.
\newblock \showarticletitle{Learning part-based templates from large
  collections of 3{D} shapes}.
\newblock \bibinfo{journal}{\emph{ACM Transactions on Graphics}}
  \bibinfo{volume}{32}, \bibinfo{number}{4} (\bibinfo{year}{2013}),
  \bibinfo{pages}{70:1--70:12}.
\newblock


\bibitem[\protect\citeauthoryear{Kim, Lipman, and Funkhouser}{Kim
  et~al\mbox{.}}{2011}]%
        {kim2011blended}
\bibfield{author}{\bibinfo{person}{Vladimir~G Kim}, \bibinfo{person}{Yaron
  Lipman}, {and} \bibinfo{person}{Thomas Funkhouser}.}
  \bibinfo{year}{2011}\natexlab{}.
\newblock \showarticletitle{Blended intrinsic maps}. In
  \bibinfo{booktitle}{\emph{ACM Transactions on Graphics}},
  Vol.~\bibinfo{volume}{30}. \bibinfo{pages}{79}.
\newblock


\bibitem[\protect\citeauthoryear{Kim, Lim, Ahn, and Kim}{Kim
  et~al\mbox{.}}{2016}]%
        {Kim2016SimultaneousSE}
\bibfield{author}{\bibinfo{person}{Youngji Kim}, \bibinfo{person}{Hwasup Lim},
  \bibinfo{person}{Sang~Chul Ahn}, {and} \bibinfo{person}{Ayoung Kim}.}
  \bibinfo{year}{2016}\natexlab{}.
\newblock \showarticletitle{Simultaneous segmentation, estimation and analysis
  of articulated motion from dense point cloud sequence}. In
  \bibinfo{booktitle}{\emph{Proc. IROS}}.
\newblock


\bibitem[\protect\citeauthoryear{Klokov and Lempitsky}{Klokov and
  Lempitsky}{2017}]%
        {klokov2017escape}
\bibfield{author}{\bibinfo{person}{Roman Klokov} {and} \bibinfo{person}{Victor
  Lempitsky}.} \bibinfo{year}{2017}\natexlab{}.
\newblock \showarticletitle{Escape from Cells: Deep {Kd-Networks} for The
  Recognition of 3{D} Point Cloud Models}. In \bibinfo{booktitle}{\emph{Proc.
  ICCV}}.
\newblock


\bibitem[\protect\citeauthoryear{Kr{\"a}henb{\"u}hl and
  Koltun}{Kr{\"a}henb{\"u}hl and Koltun}{2013}]%
        {krahenbuhl2013parameter}
\bibfield{author}{\bibinfo{person}{Philipp Kr{\"a}henb{\"u}hl} {and}
  \bibinfo{person}{Vladlen Koltun}.} \bibinfo{year}{2013}\natexlab{}.
\newblock \showarticletitle{Parameter learning and convergent inference for
  dense random fields}. In \bibinfo{booktitle}{\emph{Proc. ICML}}.
\newblock


\bibitem[\protect\citeauthoryear{Kuhn}{Kuhn}{1955}]%
        {kuhn1955hungarian}
\bibfield{author}{\bibinfo{person}{Harold~W Kuhn}.}
  \bibinfo{year}{1955}\natexlab{}.
\newblock \showarticletitle{The Hungarian method for the assignment problem}.
\newblock \bibinfo{journal}{\emph{Naval Research Logistics (NRL)}}
  \bibinfo{volume}{2}, \bibinfo{number}{1-2} (\bibinfo{year}{1955}),
  \bibinfo{pages}{83--97}.
\newblock


\bibitem[\protect\citeauthoryear{Li, Wan, Li, Sharf, Xu, and Chen}{Li
  et~al\mbox{.}}{2016}]%
        {Li:2016:MFU}
\bibfield{author}{\bibinfo{person}{Hao Li}, \bibinfo{person}{Guowei Wan},
  \bibinfo{person}{Honghua Li}, \bibinfo{person}{Andrei Sharf},
  \bibinfo{person}{Kai Xu}, {and} \bibinfo{person}{Baoquan Chen}.}
  \bibinfo{year}{2016}\natexlab{}.
\newblock \showarticletitle{Mobility Fitting Using 4D RANSAC}.
\newblock \bibinfo{journal}{\emph{Computer Graphics Forum}}
  \bibinfo{volume}{35}, \bibinfo{number}{5} (\bibinfo{year}{2016}).
\newblock


\bibitem[\protect\citeauthoryear{Liu, Qi, and Guibas}{Liu
  et~al\mbox{.}}{2018}]%
        {liu20183dflow}
\bibfield{author}{\bibinfo{person}{Xingyu Liu}, \bibinfo{person}{Charles~R Qi},
  {and} \bibinfo{person}{Leonidas~J Guibas}.} \bibinfo{year}{2018}\natexlab{}.
\newblock \showarticletitle{Learning Scene Flow in 3D Point Clouds}.
\newblock \bibinfo{journal}{\emph{arXiv preprint}} (\bibinfo{year}{2018}).
\newblock


\bibitem[\protect\citeauthoryear{Maron, Galun, Aigerman, Trope, Dym, Yumer,
  Kim, and Lipman}{Maron et~al\mbox{.}}{2017}]%
        {Maron2017CNN}
\bibfield{author}{\bibinfo{person}{Haggai Maron}, \bibinfo{person}{Meirav
  Galun}, \bibinfo{person}{Noam Aigerman}, \bibinfo{person}{Miri Trope},
  \bibinfo{person}{Nadav Dym}, \bibinfo{person}{Ersin Yumer},
  \bibinfo{person}{Vladimir~G. Kim}, {and} \bibinfo{person}{Yaron Lipman}.}
  \bibinfo{year}{2017}\natexlab{}.
\newblock \showarticletitle{Convolutional Neural Networks on Surfaces via
  Seamless Toric Covers}.
\newblock \bibinfo{journal}{\emph{ACM Transactions on Graphics}}
  \bibinfo{volume}{36}, \bibinfo{number}{4} (\bibinfo{year}{2017}).
\newblock


\bibitem[\protect\citeauthoryear{Masci, Boscaini, Bronstein, and
  Vandergheynst}{Masci et~al\mbox{.}}{2015}]%
        {masci2015geodesic}
\bibfield{author}{\bibinfo{person}{Jonathan Masci}, \bibinfo{person}{Davide
  Boscaini}, \bibinfo{person}{Michael Bronstein}, {and} \bibinfo{person}{Pierre
  Vandergheynst}.} \bibinfo{year}{2015}\natexlab{}.
\newblock \showarticletitle{Geodesic convolutional neural networks on
  {R}iemannian manifolds}. In \bibinfo{booktitle}{\emph{Proc. ICCV Workshops}}.
\newblock


\bibitem[\protect\citeauthoryear{Maturana and Scherer}{Maturana and
  Scherer}{2015}]%
        {maturana2015voxnets}
\bibfield{author}{\bibinfo{person}{Daniel Maturana} {and}
  \bibinfo{person}{Sebastian Scherer}.} \bibinfo{year}{2015}\natexlab{}.
\newblock \showarticletitle{{3D} Convolutional Neural Networks for Landing Zone
  Detection from {LiDAR}}. In \bibinfo{booktitle}{\emph{Proc. ICRA}}.
\newblock


\bibitem[\protect\citeauthoryear{Mehta, Sridhar, Sotnychenko, Rhodin, Shafiei,
  Seidel, Xu, Casas, and Theobalt}{Mehta et~al\mbox{.}}{2017}]%
        {VNect2017}
\bibfield{author}{\bibinfo{person}{Dushyant Mehta}, \bibinfo{person}{Srinath
  Sridhar}, \bibinfo{person}{Oleksandr Sotnychenko}, \bibinfo{person}{Helge
  Rhodin}, \bibinfo{person}{Mohammad Shafiei}, \bibinfo{person}{Hans-Peter
  Seidel}, \bibinfo{person}{Weipeng Xu}, \bibinfo{person}{Dan Casas}, {and}
  \bibinfo{person}{Christian Theobalt}.} \bibinfo{year}{2017}\natexlab{}.
\newblock \showarticletitle{VNect: Real-time 3D Human Pose Estimation with a
  Single RGB Camera}.
\newblock \bibinfo{journal}{\emph{ACM Transactions on Graphics}}
  \bibinfo{volume}{36}, \bibinfo{number}{4} (\bibinfo{year}{2017}), 14.
\newblock


\bibitem[\protect\citeauthoryear{Mitra, Guibas, and Pauly}{Mitra
  et~al\mbox{.}}{2006}]%
        {Mitra:2006:PAS}
\bibfield{author}{\bibinfo{person}{Niloy~J. Mitra},
  \bibinfo{person}{Leonidas~J. Guibas}, {and} \bibinfo{person}{Mark Pauly}.}
  \bibinfo{year}{2006}\natexlab{}.
\newblock \showarticletitle{Partial and Approximate Symmetry Detection for 3D
  Geometry}.
\newblock \bibinfo{journal}{\emph{ACM Transactions on Graphics}}
  \bibinfo{volume}{25}, \bibinfo{number}{3} (\bibinfo{year}{2006}).
\newblock


\bibitem[\protect\citeauthoryear{Monti, Boscaini, Masci, Rodola, Svoboda, and
  Bronstein.}{Monti et~al\mbox{.}}{2017}]%
        {Monti2017}
\bibfield{author}{\bibinfo{person}{Federico Monti}, \bibinfo{person}{Davide
  Boscaini}, \bibinfo{person}{Jonathan Masci}, \bibinfo{person}{Emanuele
  Rodola}, \bibinfo{person}{Jan Svoboda}, {and} \bibinfo{person}{Michael~M.
  Bronstein.}} \bibinfo{year}{2017}\natexlab{}.
\newblock \showarticletitle{Geometric deep learning on graphs and manifolds
  using mixture model {CNN}s}. In \bibinfo{booktitle}{\emph{Proc. CVPR}}.
\newblock


\bibitem[\protect\citeauthoryear{Newell, Yang, and Deng}{Newell
  et~al\mbox{.}}{2016}]%
        {NewellYD16}
\bibfield{author}{\bibinfo{person}{Alejandro Newell}, \bibinfo{person}{Kaiyu
  Yang}, {and} \bibinfo{person}{Jia Deng}.} \bibinfo{year}{2016}\natexlab{}.
\newblock \showarticletitle{Stacked Hourglass Networks for Human Pose
  Estimation}. In \bibinfo{booktitle}{\emph{Proc. ECCV}}.
\newblock


\bibitem[\protect\citeauthoryear{Palmer}{Palmer}{1977}]%
        {Palmer:1977}
\bibfield{author}{\bibinfo{person}{Stephen Palmer}.}
  \bibinfo{year}{1977}\natexlab{}.
\newblock \showarticletitle{Hierarchical structure in perceptual
  representation}.
\newblock \bibinfo{journal}{\emph{Cognitive Psychology}} \bibinfo{volume}{9},
  \bibinfo{number}{4} (\bibinfo{year}{1977}), \bibinfo{pages}{441--474}.
\newblock


\bibitem[\protect\citeauthoryear{Pillai, Walter, and Teller}{Pillai
  et~al\mbox{.}}{2014}]%
        {Pillai2014articulatedmotions}
\bibfield{author}{\bibinfo{person}{Sudeep Pillai}, \bibinfo{person}{Matthew~R.
  Walter}, {and} \bibinfo{person}{Seth~J. Teller}.}
  \bibinfo{year}{2014}\natexlab{}.
\newblock \showarticletitle{Learning Articulated Motions From Visual
  Demonstration}. In \bibinfo{booktitle}{\emph{Robotics: Science and Systems}}.
\newblock


\bibitem[\protect\citeauthoryear{Pirk, Krs, Hu, Rajasekaran, Kang, Yoshiyasu,
  Benes, and Guibas}{Pirk et~al\mbox{.}}{2017}]%
        {Pirk:2017}
\bibfield{author}{\bibinfo{person}{S\"{o}ren Pirk}, \bibinfo{person}{Vojtech
  Krs}, \bibinfo{person}{Kaimo Hu}, \bibinfo{person}{Suren~Deepak Rajasekaran},
  \bibinfo{person}{Hao Kang}, \bibinfo{person}{Yusuke Yoshiyasu},
  \bibinfo{person}{Bedrich Benes}, {and} \bibinfo{person}{Leonidas~J. Guibas}.}
  \bibinfo{year}{2017}\natexlab{}.
\newblock \showarticletitle{Understanding and Exploiting Object Interaction
  Landscapes}.
\newblock \bibinfo{journal}{\emph{ACM Transactions on Graphics}}
  \bibinfo{volume}{36}, \bibinfo{number}{3} (\bibinfo{year}{2017}).
\newblock


\bibitem[\protect\citeauthoryear{Qi, Su, Mo, and Guibas}{Qi
  et~al\mbox{.}}{2017a}]%
        {qi2017pointnet}
\bibfield{author}{\bibinfo{person}{Charles~R Qi}, \bibinfo{person}{Hao Su},
  \bibinfo{person}{Kaichun Mo}, {and} \bibinfo{person}{Leonidas~J Guibas}.}
  \bibinfo{year}{2017}\natexlab{a}.
\newblock \showarticletitle{{PointNet}: Deep Learning on Point Sets for {3D}
  Classification and Segmentation}. In \bibinfo{booktitle}{\emph{Proc. CVPR}}.
\newblock


\bibitem[\protect\citeauthoryear{Qi, Yi, Su, and Guibas}{Qi
  et~al\mbox{.}}{2017b}]%
        {qi2017pointnetpp}
\bibfield{author}{\bibinfo{person}{Charles~R. Qi}, \bibinfo{person}{Li Yi},
  \bibinfo{person}{Hao Su}, {and} \bibinfo{person}{Leonidas Guibas}.}
  \bibinfo{year}{2017}\natexlab{b}.
\newblock \showarticletitle{{PointNet++}: Deep Hierarchical Feature Learning on
  Point Sets in a Metric Space}. In \bibinfo{booktitle}{\emph{Proc. NIPS}}.
\newblock


\bibitem[\protect\citeauthoryear{Qi, Yi, Su, and Guibas}{Qi
  et~al\mbox{.}}{2017c}]%
        {qi2017pointnetplusplus}
\bibfield{author}{\bibinfo{person}{Charles~R Qi}, \bibinfo{person}{Li Yi},
  \bibinfo{person}{Hao Su}, {and} \bibinfo{person}{Leonidas~J Guibas}.}
  \bibinfo{year}{2017}\natexlab{c}.
\newblock \showarticletitle{PointNet++: Deep Hierarchical Feature Learning on
  Point Sets in a Metric Space}. In \bibinfo{booktitle}{\emph{Proc. NIPS}}.
\newblock


\bibitem[\protect\citeauthoryear{Quiroga, Brox, Devernay, and Crowley}{Quiroga
  et~al\mbox{.}}{2014}]%
        {quiroga2014}
\bibfield{author}{\bibinfo{person}{Julian Quiroga}, \bibinfo{person}{Thomas
  Brox}, \bibinfo{person}{Fr{\'e}d{\'e}ric Devernay}, {and}
  \bibinfo{person}{James~L. Crowley}.} \bibinfo{year}{2014}\natexlab{}.
\newblock \showarticletitle{{Dense Semi-Rigid Scene Flow Estimation from RGBD
  images}}. In \bibinfo{booktitle}{\emph{Proc. ECCV}}.
\newblock


\bibitem[\protect\citeauthoryear{Riegler, Ulusoys, and Geiger}{Riegler
  et~al\mbox{.}}{2017}]%
        {riegler2017octnet}
\bibfield{author}{\bibinfo{person}{Gernot Riegler}, \bibinfo{person}{Ali~Osman
  Ulusoys}, {and} \bibinfo{person}{Andreas Geiger}.}
  \bibinfo{year}{2017}\natexlab{}.
\newblock \showarticletitle{Octnet: Learning deep 3{D} representations at high
  resolutions}. In \bibinfo{booktitle}{\emph{Proc. CVPR}}.
\newblock


\bibitem[\protect\citeauthoryear{Romera-Paredes and Torr}{Romera-Paredes and
  Torr}{2016}]%
        {romera2016recurrent}
\bibfield{author}{\bibinfo{person}{Bernardino Romera-Paredes} {and}
  \bibinfo{person}{Philip Hilaire~Sean Torr}.} \bibinfo{year}{2016}\natexlab{}.
\newblock \showarticletitle{Recurrent instance segmentation}. In
  \bibinfo{booktitle}{\emph{Proc. ECCV}}.
\newblock


\bibitem[\protect\citeauthoryear{Rosenblatt}{Rosenblatt}{1961}]%
        {rosenblatt1961principles}
\bibfield{author}{\bibinfo{person}{Frank Rosenblatt}.}
  \bibinfo{year}{1961}\natexlab{}.
\newblock \bibinfo{booktitle}{\emph{Principles of neurodynamics. perceptrons
  and the theory of brain mechanisms}}.
\newblock \bibinfo{type}{{T}echnical {R}eport}. \bibinfo{institution}{CORNELL
  AERONAUTICAL LAB INC BUFFALO NY}.
\newblock


\bibitem[\protect\citeauthoryear{Shao, Shah, Dwaracherla, and Bohg}{Shao
  et~al\mbox{.}}{2018}]%
        {shao2018motion}
\bibfield{author}{\bibinfo{person}{Lin Shao}, \bibinfo{person}{Parth Shah},
  \bibinfo{person}{Vikranth Dwaracherla}, {and} \bibinfo{person}{Jeannette
  Bohg}.} \bibinfo{year}{2018}\natexlab{}.
\newblock \showarticletitle{Motion-based Object Segmentation based on Dense
  RGB-D Scene Flow}.
\newblock \bibinfo{journal}{\emph{arXiv preprint arXiv:1804.05195}}
  (\bibinfo{year}{2018}).
\newblock


\bibitem[\protect\citeauthoryear{Shotton, Girshick, Fitzgibbon, Sharp, Cook,
  Finocchio, Moore, Kohli, Criminisi, Kipman, and Blake}{Shotton
  et~al\mbox{.}}{2013}]%
        {Shotton:2013:EHP}
\bibfield{author}{\bibinfo{person}{Jamie Shotton}, \bibinfo{person}{Ross
  Girshick}, \bibinfo{person}{Andrew Fitzgibbon}, \bibinfo{person}{Toby Sharp},
  \bibinfo{person}{Mat Cook}, \bibinfo{person}{Mark Finocchio},
  \bibinfo{person}{Richard Moore}, \bibinfo{person}{Pushmeet Kohli},
  \bibinfo{person}{Antonio Criminisi}, \bibinfo{person}{Alex Kipman}, {and}
  \bibinfo{person}{Andrew Blake}.} \bibinfo{year}{2013}\natexlab{}.
\newblock \showarticletitle{Efficient Human Pose Estimation from Single Depth
  Images}.
\newblock \bibinfo{journal}{\emph{IEEE Transactions on Pattern Analysis and
  Machine Intelligence}} \bibinfo{volume}{35}, \bibinfo{number}{12}
  (\bibinfo{year}{2013}).
\newblock


\bibitem[\protect\citeauthoryear{Sidi, van Kaick, Kleiman, Zhang, and
  Cohen-Or}{Sidi et~al\mbox{.}}{2011}]%
        {sidi11cosegmentation}
\bibfield{author}{\bibinfo{person}{Oana Sidi}, \bibinfo{person}{Oliver van
  Kaick}, \bibinfo{person}{Yanir Kleiman}, \bibinfo{person}{Hao Zhang}, {and}
  \bibinfo{person}{Daniel Cohen-Or}.} \bibinfo{year}{2011}\natexlab{}.
\newblock \showarticletitle{Unsupervised Co-Segmentation of a Set of Shapes via
  Descriptor-Space Spectral Clustering}.
\newblock \bibinfo{journal}{\emph{ACM Transactions on Graphics}}
  \bibinfo{volume}{30}, \bibinfo{number}{6} (\bibinfo{year}{2011}).
\newblock


\bibitem[\protect\citeauthoryear{Sorkine and Alexa}{Sorkine and Alexa}{2007}]%
        {Sorkine:2007:ASM}
\bibfield{author}{\bibinfo{person}{Olga Sorkine} {and} \bibinfo{person}{Marc
  Alexa}.} \bibinfo{year}{2007}\natexlab{}.
\newblock \showarticletitle{As-rigid-as-possible Surface Modeling}. In
  \bibinfo{booktitle}{\emph{Proc. SGP}}.
\newblock


\bibitem[\protect\citeauthoryear{St\"{u}ckler and Behnke}{St\"{u}ckler and
  Behnke}{2015}]%
        {Stuckler:2015:EDR}
\bibfield{author}{\bibinfo{person}{J\"{o}rg St\"{u}ckler} {and}
  \bibinfo{person}{Sven Behnke}.} \bibinfo{year}{2015}\natexlab{}.
\newblock \showarticletitle{Efficient Dense Rigid-Body Motion Segmentation and
  Estimation in RGB-D Video}.
\newblock \bibinfo{journal}{\emph{International Journal of Computer Vision}}
  \bibinfo{volume}{113}, \bibinfo{number}{3} (\bibinfo{year}{2015}).
\newblock


\bibitem[\protect\citeauthoryear{Su, Jampani, Sun, Maji, Kalogerakis, Yang, and
  Kautz}{Su et~al\mbox{.}}{2018}]%
        {su18splatnet}
\bibfield{author}{\bibinfo{person}{Hang Su}, \bibinfo{person}{Varun Jampani},
  \bibinfo{person}{Deqing Sun}, \bibinfo{person}{Subhransu Maji},
  \bibinfo{person}{Evangelos Kalogerakis}, \bibinfo{person}{Ming-Hsuan Yang},
  {and} \bibinfo{person}{Jan Kautz}.} \bibinfo{year}{2018}\natexlab{}.
\newblock \showarticletitle{SPLATNet: Sparse Lattice Networks for Point Cloud
  Processing}. In \bibinfo{booktitle}{\emph{Proc. CVPR}}.
\newblock


\bibitem[\protect\citeauthoryear{Sumner, Schmid, and Pauly}{Sumner
  et~al\mbox{.}}{2007}]%
        {Sumner:2007}
\bibfield{author}{\bibinfo{person}{Robert~W. Sumner}, \bibinfo{person}{Johannes
  Schmid}, {and} \bibinfo{person}{Mark Pauly}.}
  \bibinfo{year}{2007}\natexlab{}.
\newblock \showarticletitle{Embedded Deformation for Shape Manipulation}.
\newblock \bibinfo{journal}{\emph{ACM Transactions on Graphics}}
  \bibinfo{volume}{26}, \bibinfo{number}{3} (\bibinfo{date}{July}
  \bibinfo{year}{2007}).
\newblock


\bibitem[\protect\citeauthoryear{Tom{\`{e}}, Russell, and Agapito}{Tom{\`{e}}
  et~al\mbox{.}}{2017}]%
        {TomeRA17}
\bibfield{author}{\bibinfo{person}{Denis Tom{\`{e}}}, \bibinfo{person}{Chris
  Russell}, {and} \bibinfo{person}{Lourdes Agapito}.}
  \bibinfo{year}{2017}\natexlab{}.
\newblock \showarticletitle{Lifting from the Deep: Convolutional 3D Pose
  Estimation from a Single Image}. In \bibinfo{booktitle}{\emph{Proc. CVPR}}.
\newblock


\bibitem[\protect\citeauthoryear{Toshev and Szegedy}{Toshev and
  Szegedy}{2014}]%
        {Toshev:2014:DHP}
\bibfield{author}{\bibinfo{person}{Alexander Toshev} {and}
  \bibinfo{person}{Christian Szegedy}.} \bibinfo{year}{2014}\natexlab{}.
\newblock \showarticletitle{DeepPose: Human Pose Estimation via Deep Neural
  Networks}. In \bibinfo{booktitle}{\emph{Proc. CVPR}}.
\newblock


\bibitem[\protect\citeauthoryear{Tzionas and Gall}{Tzionas and Gall}{2016a}]%
        {Tzionas:ECCVw:2016}
\bibfield{author}{\bibinfo{person}{Dimitrios Tzionas} {and}
  \bibinfo{person}{Juergen Gall}.} \bibinfo{year}{2016}\natexlab{a}.
\newblock \showarticletitle{Reconstructing Articulated Rigged Models from RGB-D
  Videos}. In \bibinfo{booktitle}{\emph{Proc. ECCV}}.
\newblock


\bibitem[\protect\citeauthoryear{Tzionas and Gall}{Tzionas and Gall}{2016b}]%
        {tzionas2016reconstructing}
\bibfield{author}{\bibinfo{person}{Dimitrios Tzionas} {and}
  \bibinfo{person}{Juergen Gall}.} \bibinfo{year}{2016}\natexlab{b}.
\newblock \showarticletitle{Reconstructing Articulated Rigged Models from RGB-D
  Videos}. In \bibinfo{booktitle}{\emph{Proc. ECCV}}.
\newblock


\bibitem[\protect\citeauthoryear{van Kaick, Xu, Zhang, Wang, Sun, Shamir, and
  Cohen-Or}{van Kaick et~al\mbox{.}}{2013}]%
        {vanKaick:2013:CAS}
\bibfield{author}{\bibinfo{person}{Oliver van Kaick}, \bibinfo{person}{Kai Xu},
  \bibinfo{person}{Hao Zhang}, \bibinfo{person}{Yanzhen Wang},
  \bibinfo{person}{Shuyang Sun}, \bibinfo{person}{Ariel Shamir}, {and}
  \bibinfo{person}{Daniel Cohen-Or}.} \bibinfo{year}{2013}\natexlab{}.
\newblock \showarticletitle{Co-hierarchical Analysis of Shape Structures}.
\newblock \bibinfo{journal}{\emph{ACM Transactions on Graphics.}}
  \bibinfo{volume}{32}, \bibinfo{number}{4} (\bibinfo{year}{2013}).
\newblock


\bibitem[\protect\citeauthoryear{Vogel, Roth, and Schindler}{Vogel
  et~al\mbox{.}}{2014}]%
        {vogel2014view}
\bibfield{author}{\bibinfo{person}{Christoph Vogel}, \bibinfo{person}{Stefan
  Roth}, {and} \bibinfo{person}{Konrad Schindler}.}
  \bibinfo{year}{2014}\natexlab{}.
\newblock \showarticletitle{View-consistent 3D scene flow estimation over
  multiple frames}. In \bibinfo{booktitle}{\emph{European Conference on
  Computer Vision}}. Springer, \bibinfo{pages}{263--278}.
\newblock


\bibitem[\protect\citeauthoryear{Wang, Liu, Guo, Sun, and Tong}{Wang
  et~al\mbox{.}}{2017}]%
        {wang2017ocnn}
\bibfield{author}{\bibinfo{person}{Peng-Shuai Wang}, \bibinfo{person}{Yang
  Liu}, \bibinfo{person}{Yu-Xiao Guo}, \bibinfo{person}{Chun-Yu Sun}, {and}
  \bibinfo{person}{Xin Tong}.} \bibinfo{year}{2017}\natexlab{}.
\newblock \showarticletitle{{O-CNN}: Octree-based Convolutional Neural Networks
  for {3D} Shape Analysis}.
\newblock \bibinfo{journal}{\emph{ACM Transactions on Graphics}}
  \bibinfo{volume}{36}, \bibinfo{number}{4} (\bibinfo{year}{2017}).
\newblock


\bibitem[\protect\citeauthoryear{Xu, Kim, Huang, Mitra, and Kalogerakis}{Xu
  et~al\mbox{.}}{2016}]%
        {Xu:2016:DSA}
\bibfield{author}{\bibinfo{person}{Kai Xu}, \bibinfo{person}{Vladimir~G. Kim},
  \bibinfo{person}{Qixing Huang}, \bibinfo{person}{Niloy Mitra}, {and}
  \bibinfo{person}{Evangelos Kalogerakis}.} \bibinfo{year}{2016}\natexlab{}.
\newblock \showarticletitle{Data-driven Shape Analysis and Processing}. In
  \bibinfo{booktitle}{\emph{SIGGRAPH ASIA 2016 Courses}}.
\newblock


\bibitem[\protect\citeauthoryear{Yan and Xiang}{Yan and Xiang}{2016}]%
        {yan2016scene}
\bibfield{author}{\bibinfo{person}{Zike Yan} {and} \bibinfo{person}{Xuezhi
  Xiang}.} \bibinfo{year}{2016}\natexlab{}.
\newblock \showarticletitle{Scene Flow Estimation: A Survey}.
\newblock \bibinfo{journal}{\emph{arXiv preprint arXiv:1612.02590}}
  (\bibinfo{year}{2016}).
\newblock


\bibitem[\protect\citeauthoryear{Yi, Kim, Ceylan, Shen, Yan, Su, Lu, Huang,
  Sheffer, Guibas, et~al\mbox{.}}{Yi et~al\mbox{.}}{2016}]%
        {yi2016scalable}
\bibfield{author}{\bibinfo{person}{Li Yi}, \bibinfo{person}{Vladimir~G Kim},
  \bibinfo{person}{Duygu Ceylan}, \bibinfo{person}{I Shen},
  \bibinfo{person}{Mengyan Yan}, \bibinfo{person}{Hao Su},
  \bibinfo{person}{ARCewu Lu}, \bibinfo{person}{Qixing Huang},
  \bibinfo{person}{Alla Sheffer}, \bibinfo{person}{Leonidas Guibas},
  {et~al\mbox{.}}} \bibinfo{year}{2016}\natexlab{}.
\newblock \showarticletitle{A scalable active framework for region annotation
  in 3{D} shape collections}.
\newblock \bibinfo{journal}{\emph{ACM Transactions on Graphics}}
  \bibinfo{volume}{35}, \bibinfo{number}{6} (\bibinfo{year}{2016}),
  \bibinfo{pages}{210}.
\newblock


\bibitem[\protect\citeauthoryear{Yi, Su, Guo, and Guibas}{Yi
  et~al\mbox{.}}{2017}]%
        {yi2017syncspeccnn}
\bibfield{author}{\bibinfo{person}{Li Yi}, \bibinfo{person}{Hao Su},
  \bibinfo{person}{Xingwen Guo}, {and} \bibinfo{person}{Leonidas Guibas}.}
  \bibinfo{year}{2017}\natexlab{}.
\newblock \showarticletitle{{SyncSpecCNN}: Synchronized spectral {CNN} for 3{D}
  shape segmentation}. In \bibinfo{booktitle}{\emph{Proc. CVPR}}.
\newblock


\bibitem[\protect\citeauthoryear{Yuan, Li, Xu, Chen, and Huang}{Yuan
  et~al\mbox{.}}{2016a}]%
        {Qing2016}
\bibfield{author}{\bibinfo{person}{Qing Yuan}, \bibinfo{person}{Guiqing Li},
  \bibinfo{person}{Kai Xu}, \bibinfo{person}{Xudong Chen}, {and}
  \bibinfo{person}{Hui Huang}.} \bibinfo{year}{2016}\natexlab{a}.
\newblock \showarticletitle{{Space-Time Co-Segmentation of Articulated Point
  Cloud Sequences}}.
\newblock \bibinfo{journal}{\emph{Computer Graphics Forum}}
  \bibinfo{volume}{35}, \bibinfo{number}{2} (\bibinfo{year}{2016}).
\newblock


\bibitem[\protect\citeauthoryear{Yuan, Li, Xu, Chen, and Huang}{Yuan
  et~al\mbox{.}}{2016b}]%
        {yuan2016space}
\bibfield{author}{\bibinfo{person}{Qing Yuan}, \bibinfo{person}{Guiqing Li},
  \bibinfo{person}{Kai Xu}, \bibinfo{person}{Xudong Chen}, {and}
  \bibinfo{person}{Hui Huang}.} \bibinfo{year}{2016}\natexlab{b}.
\newblock \showarticletitle{Space-Time Co-Segmentation of Articulated Point
  Cloud Sequences}.
\newblock  \bibinfo{volume}{35}, \bibinfo{number}{2} (\bibinfo{year}{2016}),
  \bibinfo{pages}{419--429}.
\newblock


\bibitem[\protect\citeauthoryear{{\v{Z}}efran and Kumar}{{\v{Z}}efran and
  Kumar}{1998}]%
        {vzefran1998interpolation}
\bibfield{author}{\bibinfo{person}{Milo{\v{s}} {\v{Z}}efran} {and}
  \bibinfo{person}{Vijay Kumar}.} \bibinfo{year}{1998}\natexlab{}.
\newblock \showarticletitle{Interpolation schemes for rigid body motions}.
\newblock \bibinfo{journal}{\emph{Computer-Aided Design}} \bibinfo{volume}{30},
  \bibinfo{number}{3} (\bibinfo{year}{1998}), \bibinfo{pages}{179--189}.
\newblock


\bibitem[\protect\citeauthoryear{Zeng, Song, Nie{\ss}ner, Fisher, Xiao, and
  Funkhouser}{Zeng et~al\mbox{.}}{2017}]%
        {zeng20163dmatch}
\bibfield{author}{\bibinfo{person}{Andy Zeng}, \bibinfo{person}{Shuran Song},
  \bibinfo{person}{Matthias Nie{\ss}ner}, \bibinfo{person}{Matthew Fisher},
  \bibinfo{person}{Jianxiong Xiao}, {and} \bibinfo{person}{Thomas Funkhouser}.}
  \bibinfo{year}{2017}\natexlab{}.
\newblock \showarticletitle{3DMatch: Learning Local Geometric Descriptors from
  RGB-D Reconstructions}. In \bibinfo{booktitle}{\emph{Proc. CVPR}}.
\newblock


\end{thebibliography}

\pagebreak
\section*{Supplementary Material}
\label{sec:supp}
This document provides a list of supplemental materials that accompany
this paper.
\begin{enumerate}
    \item \textbf{Review of PointNet, PointNet++ and Recurrent Net} - We briefly review these three basic network building blocks which are frequently used in our framework.
    \item \textbf{Training Dataset} -  We show a few shape pairs in our training set to visualize how the dataset looks like.
    \item \textbf{Test Datasets and Per-Category Motion Segmentation} - We provide a detailed evaluation of our motion segmentation on a per category basis, along with which we also list a detailed statistics for our various test datasets, including both synthetic and real ones.
    \item \textbf{Network Architecture} -  We explain the structural details for various network blocks we used.
    \item \textbf{Other Implementation Details} - We provide more implementation details and discuss some of our design choices.
\end{enumerate}

\subsection{Review of PointNet, PointNet++ and Recurrent Net}
\subsubsection{PointNet.}
Given an unordered point set represented by   feature vectors $\{\bx_1, \bx_2, ..., \bx_N\}$  of  its point samples, a PointNet \cite{qi2017pointnet} maps the input point set into a single output feature vector $\bu$ by aggregating information from all input feature vectors in an order-invariant manner:

\begin{equation}
\bu = g(\bx_1, \bx_2, ..., \bx_N) = t^{(2)} \Big( \max\limits_{i=1...N}(t^{(1)} (\bx_i) \Big)
\end{equation}
where $t^{(1)},t^{(2)}$ are learned non-linear transformations in the form of Multi-Layer Perceptrons (MLPs \cite{rosenblatt1961principles}), $N$ is the number of input points. The max function used for aggregation ensures invariance to the point order in the input point sets. We note that a PointNet can approximate any underlying function on the input point set.

\subsubsection{PointNet++} 
\cite{qi2017pointnetplusplus} is an extension of PointNet where the aggregation operates within local neighborhoods to output localized representations. Specifically, instead of using a single max pooling operation to aggregate the whole point set, aggregation is performed within a local grouping region around a point $i$ to extract a representation $\bu_i$:
\begin{equation}
\bu_i = h(\bx_1, \bx_2, ..., \bx_N) = t^{(2)} \Big( \max\limits_{j \in B(i)}(t^{(1)} (\bx_j) \Big)
\end{equation}
where $t^{(1)},t^{(2)}$ are learned MLP transformations, $B(i)$ is a local sampled neighborhood around point $i$ (Euclidean ball). By stacking multiple sampling and grouping layers in increasingly larger neighborhoods, the aggregation captures multi-scale context around each point hierarchically.

\subsubsection{Recurrent nets.}
Recurrent net layers process input feature representations in a sequence of steps. They contain an internal hidden state which stores information about previous inputs in the sequence, while the output representation is function of both the internal state and the input representation. Their state at step $t$  is updated as follows:
\begin{equation}
\bh_t = f_{\bW}( \bh_{t-1}, \bz) 
\label{eqn:hidden_state}
\end{equation}
where $\bh_{t-1}$ is the hidden state vector of the previous step (for the first step, it is set to the zero vector), $\bz$ is the input to the net at step $t$, $f_\bW(\cdot)$ denotes some non-linear function with learnable weights $\bW$. The output of a Recurrent net layer is similarly a function of the hidden state:
\begin{equation}
\bu_t = g_\bW(\bh_t) \\
\label{eqn:output_RNN_state}
\end{equation}
where $g_\bW(\cdot)$ is function with learnable weights $\bW$.

\subsection{Training Dataset}
\begin{figure}
    \centering
    \includegraphics[width=\linewidth]{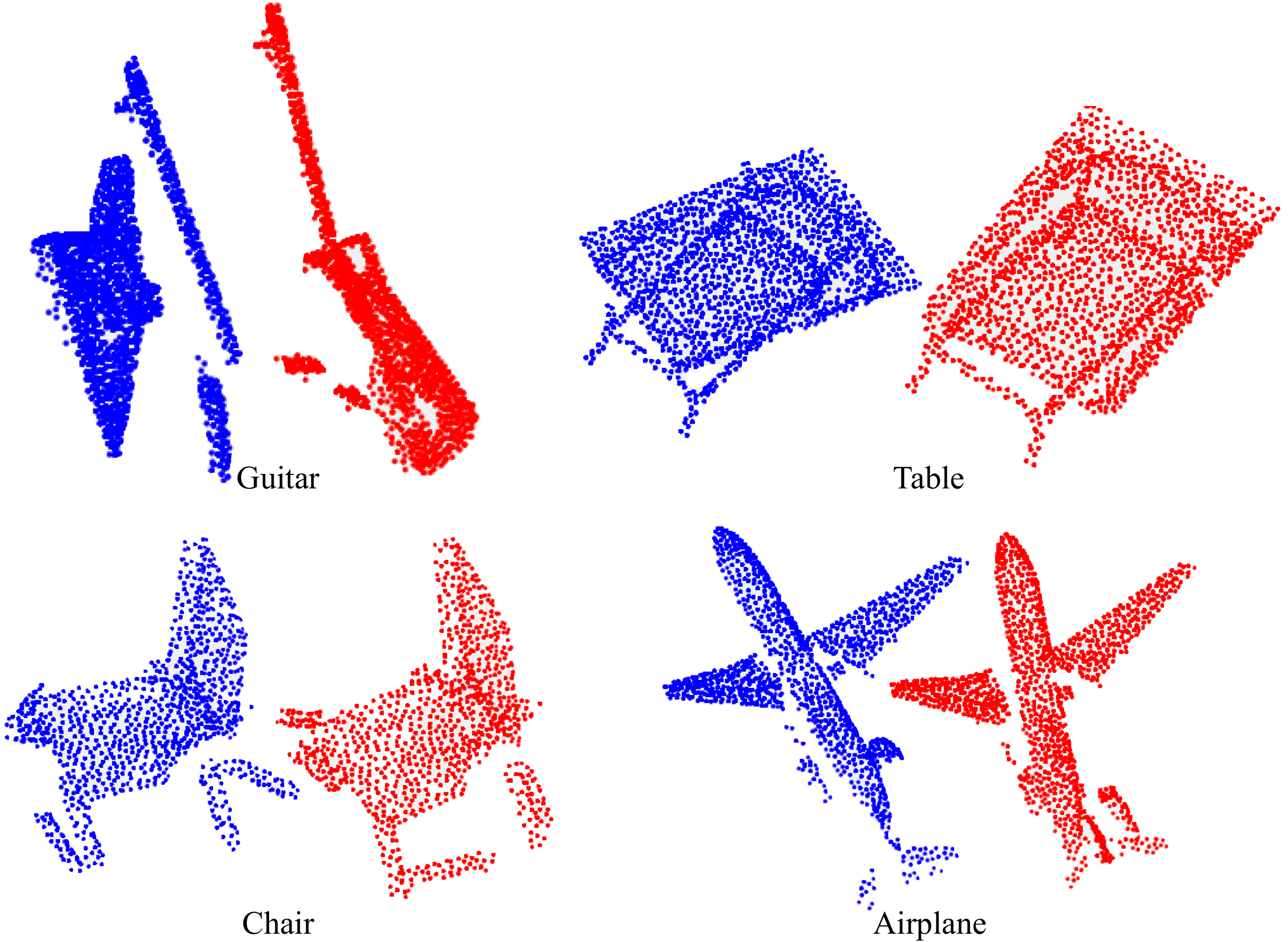}
    \caption{Visualization of the training set. Shapes are represented as point clouds. Each pair of shapes include one colored blue and one colored red.}
    \label{fig:train_set}
\end{figure}

\rev{Our training dataset contains shapes from 16 categories man-made objects. The 16 categories include airplane, earphone, cap, motorbike, bag, mug, laptop, table, guitar, knife, rocket, lamp, chair, pistol, car, skateboard.}
We visualize some of the shape pairs in our training set in Figure~\ref{fig:train_set}. Notice the generated motions cannot be guaranteed to be realistic, which means we need to learn from fake motions but generalize to real motions in our test sets.

\subsection{Test Datasets and Per-Category Motion Segmentation}
We provide detailed segmentation evaluation on a per-category basis, both on our two synthetic test datasets \textbf{SF2F}, \textbf{SF2P} and two real test datasets \textbf{RP2P}, \textbf{RF2P}. We also provide category-wise statistics for different test sets. Table~\ref{tab:sf2f_percat_seg_eva_ri_iou} shows the results on \textbf{SF2F}. Table~\ref{tab:sf2p_percat_seg_eva_ri_iou} shows the results on \textbf{SF2P}. Table~\ref{tab:rp2p_percat_seg_eva_ri_iou} shows the results on \textbf{RP2P} and Table~\ref{tab:rf2p_percat_seg_eva_ri_iou} shows the results on \textbf{RF2P}. For categories containing intrusion parts, such like cutter, syringe and telescope, the performance of our approach will degenerate. This is because points becomes hard to be differentiated when one part goes inside another. But still we outperforms all the baseline methods by a large margin.

\begin{table*}
\centering
\footnotesize
\begin{tabular}{|c|c|c|c|c|c|c|c|c|c|c|c|c|}
\hline
Category & balance & basket & box & cabinet & cutter & desk & flipLadder & flipPhone & flipUSB & fridge & luggage & notebook\\ \hline
$\#$Pairs & 10 & 50 & 20 & 70 & 55 & 30 & 25 & 35 & 50 & 50 & 25 & 40\\\hline
SeqRANSAC & 61.7/68.1 & 48.8/61.8 & 67.8/61.1 & 42.0/48.6 & 50.2/39.7 & 43.9/47.7 & 79.8/78.4 & 70.5/65.6 & 58.4/43.9 & 53.0/58.7 & 73.6/64.7 & 65.6/55.5\\
SC & \textbf{93.4}/72.7 & 92.2/72.9 & 96.2/96.0 & 88.0/61.0 & 62.6/49.6 & 87.2/55.9 & 87.2/82.8 & 92.7/91.6 & 65.4/57.3 & 90.3/71.4 & 96.8/96.7 & \textbf{93.1}/\textbf{92.7}\\
JLC & 90.7/76.2 & 88.0/82.0 & 91.8/91.7 & 84.7/71.3 & 63.1/51.6 & 87.1/68.5 & 81.7/78.0 & 91.4/90.1 & 65.5/56.9 & 87.3/76.1 & 95.4/96.4 & 86.0/85.1\\
NRR & 76.7/64.5 & 83.7/50.3 & 90.0/87.3 & 83.1/53.6 & 64.4/42.1 & 85.5/51.0 & 70.5/56.9 & 82.6/79.3 & 56.9/39.7 & 87.5/66.5 & 90.8/88.1 & 84.7/82.2\\
Ours & 91.4/\textbf{80.2} & \textbf{95.8}/\textbf{87.8} & \textbf{96.4}/\textbf{96.1} & \textbf{92.9}/\textbf{80.4} & \textbf{66.7}/\textbf{53.9} & \textbf{91.8}/\textbf{73.9} & \textbf{88.3}/\textbf{86.3} & \textbf{93.6}/\textbf{92.8} & \textbf{65.8}/\textbf{58.8} & \textbf{95.0}/\textbf{85.0} & \textbf{97.5}/\textbf{97.5} & 91.4/90.0\\\hline
Category & nutCracker & nut & plier & scissor & slidePhone & syringe & TVBench & tap & teeterBoard & telescope & windmill & overall\\ \hline
$\#$Pairs & 45 & 35 & 25 & 65 & 25 & 55 & 35 & 35 & 20 & 25 & 50 & 875\\\hline
SeqRANSAC & 83.1/78.9 & 59.0/42.8 & 73.8/70.1 & 75.8/70.9 & 53.0/34.2 & 60.9/49.3 & 44.0/45.9 & 58.1/52.9 & 68.8/55.9 & 53.0/41.5 & 65.1/58.9 & 60.2/55.8\\
SC & 92.7/92.7 & 70.2/66.8 & 80.6/80.3 & 84.8/84.2 & 56.7/46.6 & 65.7/58.5 & 86.3/59.0 & 65.1/38.7 & 72.9/59.3 & 68.7/61.5 & 76.7/66.8 & 80.6/69.4\\
JLC & 91.5/91.4 & 68.5/62.5 & 75.1/71.9 & 75.9/71.1 & 52.9/33.8 & 60.9/49.4 & 43.9/45.7 & 57.6/52.0 & 68.7/56.2 & 53.1/41.0 & 64.8/58.6 & 74.7/67.3\\
NRR & 84.8/83.4 & 56.8/35.0 & 72.3/67.9 & 69.5/61.1 & 51.6/29.2 & 59.7/42.5 & 80.0/47.3 & 79.6/68.3 & 70.8/62.7 & 58.9/36.8 & 69.2/54.3 & 74.1/57.3\\
Ours & \textbf{93.4}/\textbf{93.3} & \textbf{72.6}/\textbf{70.1} & \textbf{82.1}/\textbf{81.6} & \textbf{85.3}/\textbf{85.0} & \textbf{58.9}/\textbf{50.8} & \textbf{68.3}/\textbf{63.1} & \textbf{91.1}/\textbf{77.2} & \textbf{86.4}/\textbf{78.2} & \textbf{74.9}/\textbf{67.8} & \textbf{70.0}/\textbf{62.4} & \textbf{82.6}/\textbf{76.3} & \textbf{83.8}/\textbf{77.3}\\\hline
\end{tabular}
\caption{Per category RI and IoU evalution on synthetic dataset \textbf{SF2F} for all competing methods. We show the segmentation score in the form of RI/IoU. Both RI and IoU measure the segmentation consistency. Higher RI and higher IoU mean better segmentation prediction.  }
\label{tab:sf2f_percat_seg_eva_ri_iou}
\end{table*}

\begin{table*}
\centering
\footnotesize
\begin{tabular}{|c|c|c|c|c|c|c|c|c|c|c|c|c|}
\hline
Category & balance & basket & box & cabinet & cutter & desk & flipLadder & flipPhone & flipUSB & fridge & luggage & notebook\\ \hline
$\#$Pairs & 10 & 50 & 20 & 70 & 55 & 30 & 25 & 35 & 50 & 50 & 25 & 40\\\hline
SeqRANSAC & 30.2/38.4 & 33.8/46.6 & 59.4/43.1 & 30.9/36.5 & 42.6/30.0 & 27.9/35.9 & 67.5/64.1 & 56.1/36.8 & 51.6/25.1 & 34.7/40.8 & 62.5/42.3 & 56.4/33.8\\
SC & 55.7/50.6 & 78.0/65.9 & 82.1/79.2 & 66.1/55.1 & 52.5/38.5 & 74.6/54.6 & 80.6/76.7 & 75.1/66.0 & 56.9/41.9 & 73.6/65.7 & 82.3/79.4 & 70.9/62.2\\
JLC & 44.7/46.5 & 70.5/66.0 & 80.6/78.4 & 62.0/55.8 & 59.6/44.8 & 60.2/51.3 & 82.0/78.9 & 74.2/65.8 & 57.4/43.4 & 68.7/63.0 & 82.5/79.0 & 68.3/60.0\\
NRR & \textbf{82.9}/\textbf{58.5} & 81.0/47.7 & 62.3/46.2 & 83.6/53.1 & \textbf{67.1}/43.7 & \textbf{86.0}/51.0 & 70.3/55.3 & 81.4/79.1 & 55.0/34.5 & 83.0/57.7 & 80.0/72.4 & \textbf{81.1}/\textbf{76.0}\\
Ours & 76.9/55.6 & \textbf{91.1}/\textbf{79.8} & \textbf{86.7}/\textbf{85.2} & \textbf{84.5}/\textbf{69.0} & 63.7/\textbf{47.8} & 75.5/\textbf{58.4} & \textbf{84.4}/\textbf{81.1} & \textbf{82.0}/\textbf{80.0} & \textbf{61.3}/\textbf{52.3} & \textbf{83.7}/\textbf{68.9} & \textbf{87.5}/\textbf{85.7} & 75.6/71.7\\\hline
Category & nutCracker & nut & plier & scissor & slidePhone & syringe & TVBench & tap & teeterBoard & telescope & windmill & overall\\ \hline
$\#$Pairs & 45 & 35 & 25 & 65 & 25 & 55 & 35 & 35 & 20 & 25 & 50 & 875\\\hline
SeqRANSAC & 64.7/45.9 & 51.9/29.1 & 57.5/35.2 & 63.1/46.8 & 51.5/26.1 & 53.4/34.0 & 32.2/37.2 & 44.3/31.5 & 50.8/36.9 & 48.6/27.4 & 49.8/41.2 & 48.2/37.6\\
SC & 71.6/58.3 & 59.8/44.5 & 62.3/49.2 & 73.1/65.0 & 54.5/36.9 & 60.2/47.3 & 68.8/55.5 & 57.1/41.3 & 66.1/53.8 & 63.3/51.7 & 61.9/54.0 & 67.0/55.6\\
JLC & 80.7/76.5 & 58.2/45.7 & 65.8/57.4 & 75.7/69.1 & 54.5/38.5 & 61.7/52.4 & 62.7/57.1 & 56.2/45.0 & 69.1/58.2 & 63.1/53.4 & 62.2/56.4 & 66.2/58.2\\
NRR & 86.8/84.8 & 58.7/38.2 & 70.5/66.7 & 68.9/59.9 & 51.6/29.2 & 63.2/48.8 & 78.7/47.3 & \textbf{69.4}/45.7 & 69.1/\textbf{59.1} & 59.6/38.1 & 67.7/49.2 & 72.7/53.9\\
Ours & \textbf{87.8}/\textbf{87.1} & \textbf{64.7}/\textbf{58.3} & \textbf{71.0}/\textbf{69.2} & \textbf{79.3}/\textbf{77.8} & \textbf{55.2}/\textbf{41.8} & \textbf{64.3}/\textbf{57.4} & \textbf{81.2}/\textbf{65.3} & 66.3/\textbf{52.0} & \textbf{70.3}/56.1 & \textbf{68.8}/\textbf{59.4} & \textbf{72.9}/\textbf{65.7} & \textbf{75.6}/\textbf{66.6}\\\hline
\end{tabular}
\caption{Per category RI and IoU evalution on synthetic dataset \textbf{SF2P} for all competing methods. We show the segmentation score in the form of RI/IoU. Both RI and IoU measure the segmentation consistency. Higher RI and higher IoU mean better segmentation prediction.  }
\label{tab:sf2p_percat_seg_eva_ri_iou}
\end{table*}

\begin{table*}
\centering
\footnotesize
\begin{tabular}{|c|c|c|c|c|c|c|c|c|c|c|c|}
\hline
Category & laptop & plier & scissor & swivelChair & tableLamp & basket & box & doll & truck & tricycle & overall\\ \hline
$\#$Pairs & 39 & 16 & 31 & 30 & 22 & 25 & 24 & 12 & 23 & 9 & 231\\\hline
SeqRANSAC & 63.5/46.6 & 61.3/43.8 & 59.0/35.8 & 51.9/47.0 & 52.2/37.2 & 39.4/49.8 & 57.1/46.0 & 72.4/40.0 & 58.1/41.3 & 59.0/33.3 & 56.7/43.0\\
SC & 87.8/87.0 & 76.8/74.7 & 75.2/68.5 & 70.1/52.6 & 71.6/61.8 & 88.8/60.6 & 86.5/81.0 & 55.6/21.3 & 82.7/78.0 & 80.7/37.8 & 79.1/67.1\\
JLC & 84.3/82.3 & 73.1/68.5 & 72.7/65.4 & 82.4/75.9 & 73.2/72.1 & 89.9/80.0 & 85.8/81.8 & 71.4/40.3 & 81.3/77.8 & 82.6/48.4 & 80.4/73.0\\
NRR & 84.0/79.7 & 76.5/74.8 & 73.6/67.1 & 84.4/76.1 & \textbf{85.7}/78.6 & 86.4/52.2 & 74.7/60.6 & 55.2/25.3 & 72.2/61.2 & 69.3/28.8 & 78.4/65.5\\
Ours & \textbf{88.2}/\textbf{87.9} & \textbf{78.4}/\textbf{77.2} & \textbf{84.5}/\textbf{83.1} & \textbf{96.1}/\textbf{94.8} & 82.5/\textbf{80.9} & \textbf{96.5}/\textbf{89.0} & \textbf{92.5}/\textbf{90.0} & \textbf{83.2}/\textbf{54.5} & \textbf{85.4}/\textbf{83.1} & \textbf{88.2}/\textbf{52.6} & \textbf{88.3}/\textbf{83.5}\\\hline
\end{tabular}
\caption{Per category RI and IoU evalution on synthetic dataset \textbf{RP2P} for all competing methods. We show the segmentation score in the form of RI/IoU. Both RI and IoU measure the segmentation consistency. Higher RI and higher IoU mean better segmentation prediction.  }
\label{tab:rp2p_percat_seg_eva_ri_iou}
\end{table*}

\begin{table*}
\centering
\footnotesize
\begin{tabular}{|c|c|c|c|c|c|c|c|c|c|c|c|}
\hline
Category & laptop & plier & scissor & swivelChair & tableLamp & basket & box & dishwasher & truck & tricycle & overall\\ \hline
$\#$Pairs & 28 & 11 & 19 & 16 & 13 & 14 & 24 & 5 & 14 & 6 & 150\\\hline
SeqRANSAC & 59.8/40.7 & 61.4/42.7 & 63.4/40.5 & 65.2/54.2 & 56.8/40.4 & 32.5/42.0 & 60.9/44.0 & 58.4/46.5 & 65.3/57.9 & 57.1/28.7 & 58.7/44.2\\
SC & 73.7/65.2 & 66.6/57.4 & 73.8/65.1 & 70.2/41.2 & 64.5/52.3 & 88.7/46.5 & 64.9/44.6 & 79.6/46.7 & 72.8/62.1 & 68.4/29.1 & 71.9/53.6 \\
JLC & 66.4/57.0 & 68.1/59.3 & 74.4/65.6 & 73.5/53.5 & 80.4/75.2 & 80.9/57.0 & 65.6/48.1 & 88.2/64.9 & 77.8/69.3 & 70.3/31.8 & 72.7/58.4\\
NRR & 76.8/68.4 & 68.5/64.6 & 75.9/69.3 & 74.5/45.3 & 59.0/43.0 & 91.7/47.8 & 63.8/42.9 & 83.3/58.7 & 74.4/62.6 & 56.3/17.5 & 72.8/54.7\\
Ours & \textbf{89.7}/\textbf{88.9} & \textbf{75.5}/\textbf{74.4} & \textbf{85.9}/\textbf{85.1} & \textbf{95.4}/\textbf{88.2} & \textbf{86.2}/\textbf{85.1} & \textbf{96.9}/\textbf{81.1} & \textbf{84.2}/\textbf{78.4} & \textbf{95.7}/\textbf{90.0} & \textbf{85.1}/\textbf{81.6} & \textbf{79.5}/\textbf{37.5} & \textbf{87.6}/\textbf{81.8}\\\hline
\end{tabular}
\caption{Per category RI and IoU evalution on synthetic dataset \textbf{RF2P} for all competing methods. We show the segmentation score in the form of RI/IoU. Both RI and IoU measure the segmentation consistency. Higher RI and higher IoU mean better segmentation prediction.  }
\label{tab:rf2p_percat_seg_eva_ri_iou}
\end{table*}

\subsection{Network Architecture}
\subsubsection{PointNet Building Blocks}
We mainly use three types of PointNet/PointNet++ variations as our building blocks, which we refer to as PointNetC, PointNetS and PointNet++S. In PointNetC, point features are extracted from a mutli-layer perceptron (MLP) first and then aggregated through a max pooling operation. We use a list to denote the number of hidden nodes in the hidden layer. For example $\text{PointNetC}(K_1,K_2,K_3)$ denotes a PointNetC whose MLP contains three hidden layers with $K_1,K_2,K_3$ hidden nodes respectively. In PointNetS, features are first globally aggregated through a PointNetC, then local features before max pooling operation are concatenated together with the global feature and fed into another MLP to get a per point feature representation. We use $\text{PointNetS}[(K_1,K_2,K_3),(K_4,K_5,K_6)]$ to denote a PointNetS with a $\text{PointNetC}(K1,K2,K3)$ sub-module followed by an MLP with three hidden layers containing $K_4$,$K_5$,$K6$ hidden nodes. In PointNet++S, points are down-sampled and grouped for several times first to capture features at different scales. Then features at coarser scales are gradually up-sampled and combined with features at finer scales, resulting in a feature representation for each initial point at the end. PointNet++S involves two basic operations: sample group operation and interpolation operation. In both operations, there will be an MLP to process the point features. For a sample group operation with $N$ sampled points, a radius of $r$, a maximum number of sampled points per group $G$, and 3 hidden layers in its MLP containing $K_1$, $K_2$, $K_3$ nodes respectively, we denote it as $SG(N,r,G,(K_1,K_2,K_3))$. For an interpolation operation with 3 hidden layers in its MLP containing $K_1$, $K_2$, $K_3$ nodes respectively, we denote it as $IP(K_1,K_2,K_3)$. Then we could denote the architecture of a PointNet++S as a list of such operations, e.g. $\text{PointNet++S}[SG(N_1,r_1,G_1,(K_{11},K_{12},K_{13})), IP(\hat{K}_1,\hat{K}_2,\hat{K}_3)]$

In our correspondence proposal module, we use a shared PointNet++S to extract point features for matching purpose, which includes the following operations in a sequential order:
\begin{itemize}
    \item $SG(256,0.2,64,64,64,128)$
    \item $SG(128,0.4,64,128,128,256)$
    \item $SG(1,none,none,256,512,1024)$
    \item $IP(256,256)$
    \item $IP(256,128)$
    \item $IP(128,128,64)$
\end{itemize}
Notice $SG(1,none,none,256,512,1024)$ denotes a global grouping operation where the center of the points is sampled.

In the flow module, we use a PairNet to encode a pairwise matrix between two sets. It first globally aggregates information along the second set through the PointNet to extract a per-point representation for each point in the first set. The point representations are then hierarchically aggregated into a higher-level representation through a PointNet++, to encode local dependencies in the first set. In this PairNet, we use a $\text{PointNetC}(32,64,128,256)$ to do the first aggregation and we use a PointNet++S with the following structure for the second aggregation:
\begin{itemize}
    \item $SG(256,0.2,32,64,64,128)$
    \item $SG(128,0.4,32,128,128,256)$
    \item $SG(1,none,none,256,512,1024)$
    \item $IP(256,256)$
    \item $IP(256,128)$
    \item $IP(128,128,64)$
\end{itemize}
To extract the correspondence mask, we use a $\text{PointNetC}(32,64,128)$ followed by an MLP with 3 hidden layers containing $(128,64,32)$ hidden nodes.

In the segmentation module, we first use a PointNet++S with the following structure for hypothesis generation:
\begin{itemize}
    \item $SG(256,0.2,64,64,64,128)$
    \item $SG(128,0.4,64,128,128,256)$
    \item $SG(1,none,none,256,512,1024)$
    \item $IP(256,256)$
    \item $IP(256,128)$
    \item $IP(128,128,64)$
\end{itemize}
Then we use a $\text{PointNetS((16,64,512),(256,64,16)}$ to extract the support matrix for the generated hypotheses. Finally in the probabilistic hypothesis selection sub-module, we use a PairNet in our recurrent unit, which contains a $\text{PointNetC}(16,64,256)$ and a PointNetS with the following structure:
\begin{itemize}
    \item $SG(1,none,none,128,32,32)$
    \item $IP(32,16)$
\end{itemize}

\subsubsection{Single Image Based 3D Reconstruction Network}
In our application of part induction from a shape-image pair, we design another neural network for the purpose of lifting the 2D product image onto 3D. Given a 2D image,
for each foreground pixel, our network predicts its x, y, z coordinates in 3D space. By constraining the generated coordinates in [0, 1], the point cloud of the input product image can be represented by a 2D image (RGB channels correspond to XYZ coordinates). Our network is composed of one encoder and three decoders. We use a pre-trained ResNet-34 as our image encoder (the pooling and fully connected layers are removed). The decoder module consists of 3 branches, each containing a series of transpose convolutional and plain convolutional layers. The tasks of the 3 decoders include depth, x-y, and mask prediction. The RGB image representing the point cloud is generated by concatenating the depth with the x-y channels, after cleaning them up with the binary mask output. A U-net skip connection is added to boost information from the encoder to the decoder. There are two stages of the training. In the first stage, we jointly train our network using the ShapeNetCore dataset \cite{chang2015shapenet}, which contains a large collection of categorized 3D CAD models with texture. In the second stage, we refine our network on articulated objects. For each category, we fine-tune our network on the dataset provided by \cite{hu2017learning}, which contains articulated objects without texture.

\vspace{-0.15cm}
\subsection{Other Implementation Details}
\subsubsection{Rigid Motion Parameterization} In our hypothesis generation sub-module of the segmentation module, we adopt a ``residual'' parametrization for the rigid motions.
To be specific, we use a $3$x$3$ matrix $\hat{\bR}_i$ and a $3$x$1$ vector $\hat{\bt}_i$, from which the rotational component  is computed as $\bR_i = \hat{\bR}_i + \bI$ followed by an SVD to project the matrix to the nearest orthogonal matrix, while the translational component is computed after applying the inferred rotation: $\bt_i = -(\bR_i-I) \cdot \bx^{(p)}_i+ \bff_i + \hat{\bt}_i$, where $I$ is the identity transformation, $\bx^{(p)}_i$ is the $i^{th}$ point position and $\bff_i$ is the corresponding flow.

At training time, instead of directly optimizing $\bR_p$ and $\bt_p$, we found optimizing ``residual'' rotation matrix and translation vector $\hat{\bR}_p$ and $\hat{\bt}_p$ achieves much better performance. The motion loss term $L_{motion}$ with $\hat{\bR}_p$ and $\hat{\bt}_p$ is as below:

\begin{align*}
L_{motion} = \lambda_d \sum\limits_{ \{p,q\} \in \mM \,\,}
\sum\limits_{ \substack{ \{p',q'\} \in \mM \\ part(p')=part(p) } }
\!\!\!\!\!\!\!\! || (\bff_{p'} - \bff_p) -( \hat{\bR}_p (p'-p) + \hat{\bt}_p) ||^2
\end{align*}
where $\bff_p$ and $\bff_{p'}$ denotes the deformation flow on point $p$ and $p'$ respectively.

\subsubsection{Graph Cuts}
To convert the soft segmentation indicator functions predicted by our segmentation module into a hard segmentation, we adopt a graph cut technique. To generate a point graph, we first convert our target point set into a KNN graph, where each point gets connected with closest $K=10$ points in Euclidean space. Assuming point $p_i$ is connected with point $p_j$, then we set the weight on the connecting edge as $\text{exp}^{-\frac{||p_i-p_j||^2}{2\sigma^2}}$, where $\sigma=0.05$ in our experiments. We use the edge weights of the point graph as pairwise terms. We then treat our soft segmentation indicator functions as probabilities and use the negative log probability as the unary term. We run the graph cut algorithm described in \cite{Boykov2001} to get the hard labeling for points, namely the segmentation we are looking for.

\begin{table}[t!]
\centering
\small
\begin{tabular}{|c|c|c|c|c|}
\hline
 & (a) & (b) & (c) & (d)  \\ \hline
SF2F & \textbf{0.0377} & 0.0495 & 0.046 &      0.044  \\
SF2P & \textbf{0.0522} &       0.0771 & 0.764 & 0.0759  \\\hline
\end{tabular}
\caption{\rev{Ablation study of our network: we evaluate the deformation flow generated by different variations (a)-(d) of our framework with a single iteration}, as explained in Sec~\ref{sec:abl}, using EPE as the metric.}
\label{tab:flow_eva_our}
\end{table}

\end{document}